\documentclass[journal]{IEEEtran}
\usepackage{xcolor,soul,framed} 
\usepackage{algorithm}
\usepackage{algpseudocode}
\usepackage{booktabs}
\colorlet{shadecolor}{yellow}
\usepackage[pdftex]{graphicx}
\graphicspath{{../pdf/}{../jpeg/}}
\DeclareGraphicsExtensions{.pdf,.jpeg,.png}
\usepackage{multirow}
\usepackage[cmex10]{amsmath}
\usepackage{array}
\newcommand{\PreserveBackslash}[1]{\let\temp=\\#1\let\\=\temp}
\newcolumntype{C}[1]{>{\PreserveBackslash\centering}p{#1}}
\newcolumntype{R}[1]{>{\PreserveBackslash\raggedleft}p{#1}}
\newcolumntype{L}[1]{>{\PreserveBackslash\raggedright}p{#1}}

\usepackage{mdwmath}
\usepackage{mdwtab}
\usepackage{eqparbox}
\usepackage{url}
\usepackage{hyperref}
\usepackage{amssymb}

\usepackage{tikz,xcolor,hyperref}
\definecolor{lime}{HTML}{A6CE39}
\DeclareRobustCommand{\orcidicon}{%
	\begin{tikzpicture}
	\draw[lime, fill=lime] (0,0)
	circle [radius=0.16]
	node[white] {{\fontfamily{qag}\selectfont \tiny ID}};	\draw[white, fill=white] (-0.0625,0.095)
	circle [radius=0.007];	\end{tikzpicture}
	\hspace{-2mm}}
\foreach \x in {A, ..., Z}{%
	\expandafter\xdef\csname orcid\x\endcsname{\noexpand\href{https://orcid.org/\csname orcidauthor\x\endcsname}{\noexpand\orcidicon}}
	}

\ifCLASSINFOpdf
\else
\fi
\begin{document}
\title{KSS-ICP: Point Cloud Registration based on Kendall Shape Space}

\author{Chenlei~Lv\orcidA{},~\IEEEmembership{Member,~IEEE,}
Weisi Lin\orcidB{},~\IEEEmembership{Fellow,~IEEE,} Baoquan Zhao\orcidC{}, ~\IEEEmembership{Member,~IEEE}
\thanks{
This research is supported by the Ministry of Education, Singapore, under its Tier-1 Fund MOE2021, RG14/21.

Chenlei~Lv, Weisi Lin and Baoquan Zhao were with the School of Computer Science and Engineering, Nanyang Technological University. The correspoding author is Weisi Lin, e-mail:(wslin@ntu.edu.sg).}}

\markboth{}%
{Shell \MakeLowercase{\textit{et al.}}: Bare Demo of IEEEtran.cls for IEEE Journals}

\maketitle

\begin{abstract}
Point cloud registration is a popular topic which has been widely used in 3D model reconstruction, location, and retrieval. In this paper, we propose a new registration method, KSS-ICP, to address the rigid registration task in Kendall shape space (KSS) with Iterative Closest Point (ICP). The KSS is a quotient space that removes influences of translations, scales, and rotations for shape feature-based analysis. Such influences can be concluded as the similarity transformations that do not change the shape feature. The point cloud representation in KSS is invariant to similarity transformations. We utilize such property to design the KSS-ICP for point cloud registration. To tackle the difficulty
to achieve the KSS representation in general, the proposed KSS-ICP formulates a practical solution that does not require complex feature analysis, data training, and optimization. With a simple implementation, KSS-ICP achieves more accurate registration from point clouds. It is robust to similarity transformation, non-uniform density, noise, and defective parts. Experiments show that KSS-ICP has better performance than the state of the art. Code\footnote{\href{https://github.com/vvvwo/KSS-ICP}{Code Link: \color{blue}vvvwo/KSS-ICP}.} and executable files\footnote{\href{https://github.com/vvvwo/KSS-ICP/tree/master/EXE}{EXE Link: \color{blue}vvvwo/KSS-ICP/tree/master/EXE}.} are made public.

\end{abstract}

\begin{IEEEkeywords}
Kendall shape space, point cloud registration.
\end{IEEEkeywords}

\IEEEpeerreviewmaketitle

\section{Introduction}

With the development of 3D scanning technology, 3D point clouds have been widely used in different applications such as autopilot~\cite{chiang2017development}\cite{bahirat2018alert}, architectural design~\cite{xue2019bim}, digital animation production~\cite{barbieri20183d}\cite{furukawa2014interactive}, bioinformatics~\cite{roy2004wide}\cite{nair20093}, and medical treatment~\cite{liao2013review}. As a fundamental process for the applications, point cloud registration has been researched for many years. The target of the registration is to find the correspondence between two point clouds or build the transformation matrix from one point cloud to another. It provides basic technical support for point cloud location~\cite{zheng2020lodonet}\cite{battiato2011robust}, reconstruction~\cite{liu2019l2g}, and detection~\cite{qin2020weakly}.

There are some influences that increase the difficulty of the registration process, including similarity transformation, non-uniform density, noise, and defective part. These influences are explained as follows. The similarity transformation is a mathematical concept that means the transformation would not change the shape feature of the corresponding object. Therefore, it is also called shape-preserving transformation, including translation, rotation, and scaling. During the scanning, it cannot be guaranteed that raw point clouds have the uniform location, rotation, and scale. The influence produced by similarity transformation for the point cloud can not be avoided. Affected by occlusions, illuminations, dust in the wild, and different scanning sources (cross-source problem~\cite{huang2019fast}), points with non-uniform density, noise and defective parts are scanned into point clouds at the same time. These disturbing factors increase the probability of error correspondence's production. Such influences bring the main challenges in point cloud registration.

\begin{figure}
  \centering
  \includegraphics[width=\linewidth]{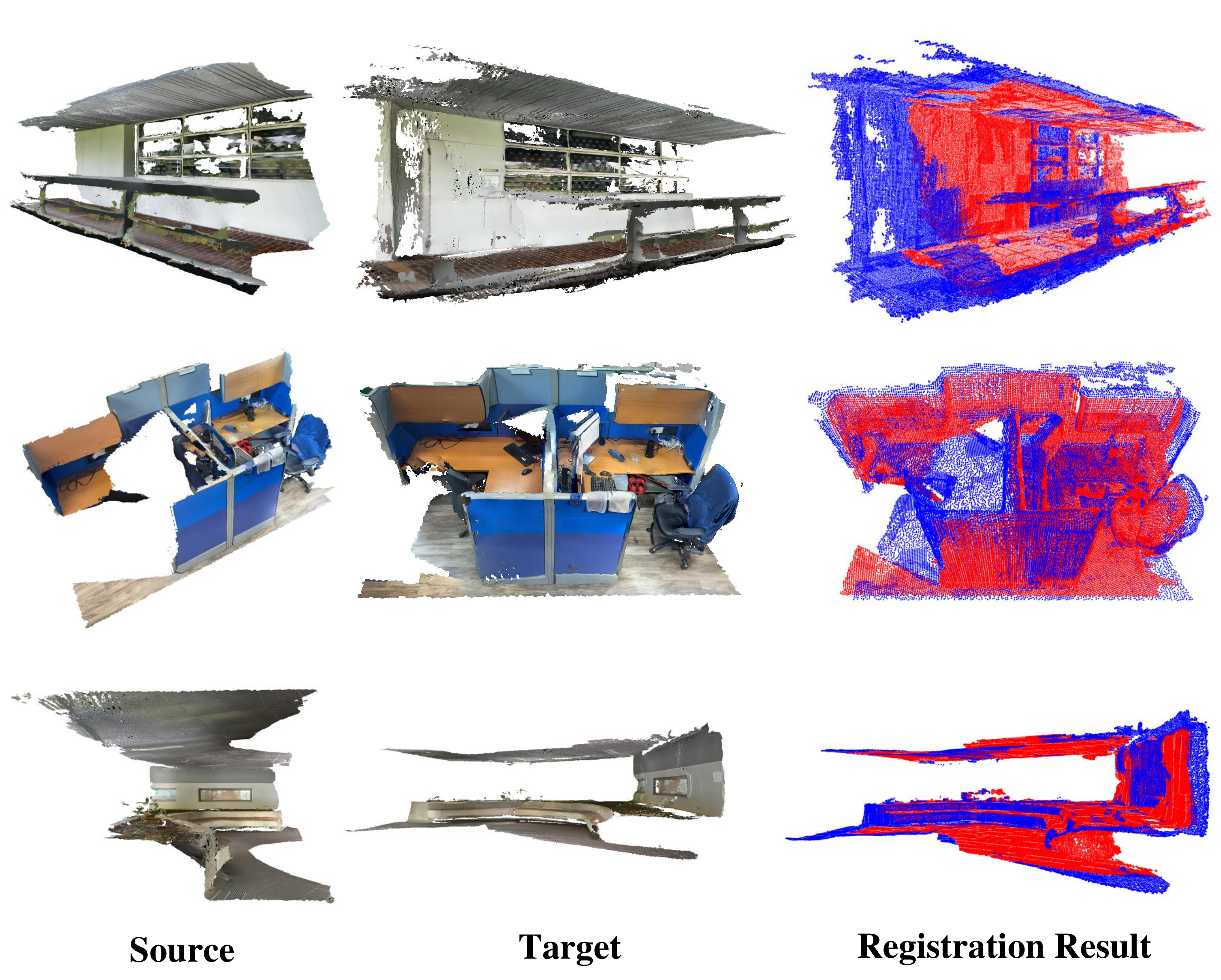}
  \caption{Instances of KSS-ICP registration results. Point cloud models are scanned by iphone12pro in Nanyang Technological University. In third column, the red point clouds represent the source point clouds; the blue point clouds are target ones. Registration results show that our method can align raw point clouds from source to target.}
  \label{f1}
\end{figure}

\begin{figure}
  \centering
  \includegraphics[width=\linewidth]{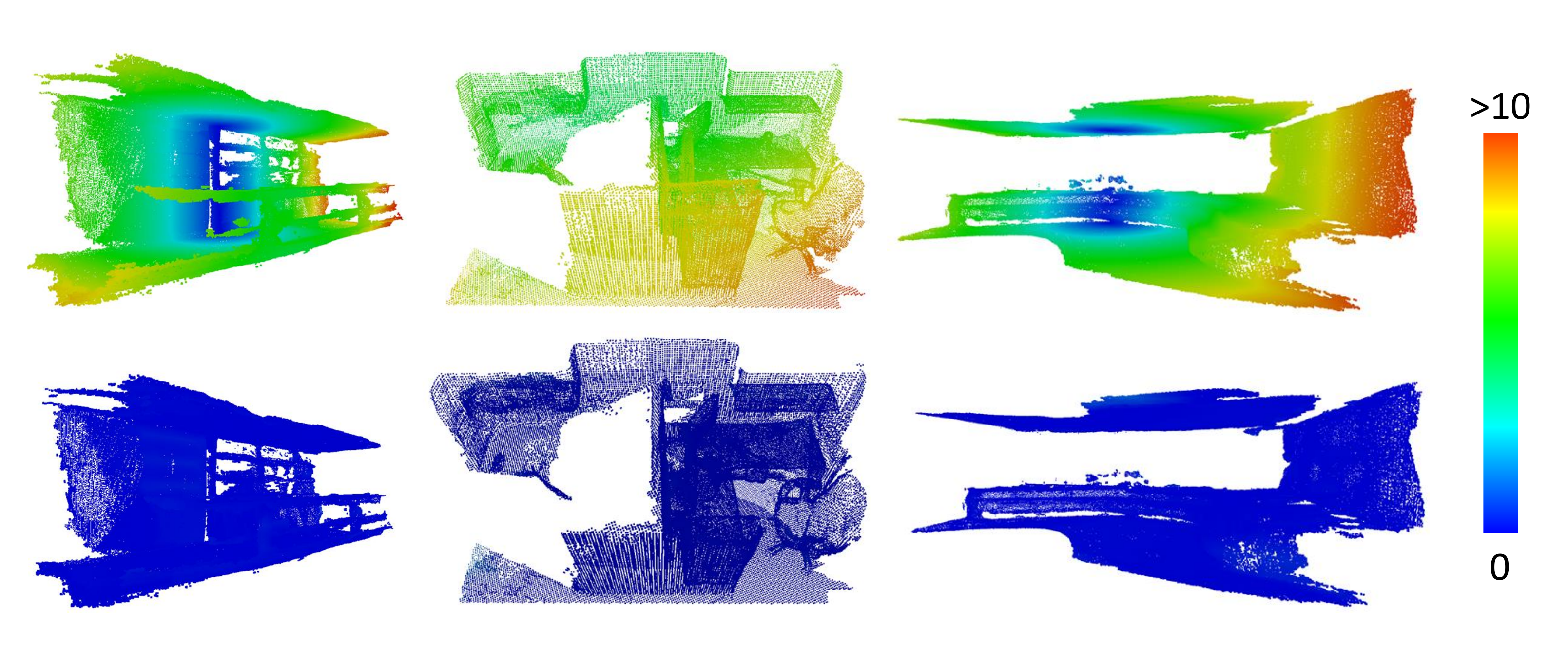}
  \caption{Color maps of corresponding errors (point distance) between source point clouds and target ones (red to blue: high to low). First row: before registration; second row: after registration. It is clear that the corresponding errors approach to zero after our registration. }
  \label{f1_2}
\end{figure}

\begin{figure*}
  \centering
  \includegraphics[width=\linewidth]{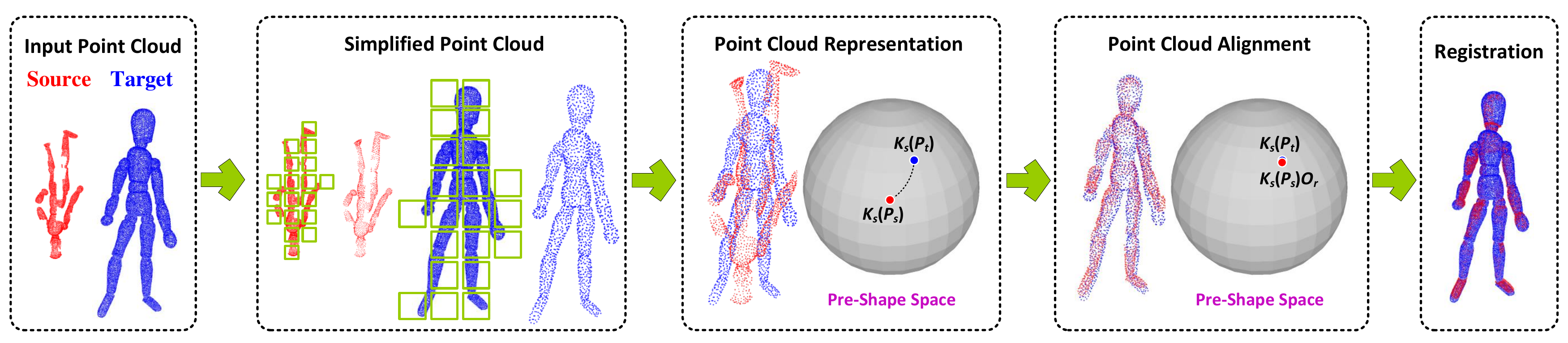}
  \caption{The pipeline of KSS-ICP. In pre-shape space, the influence of rotation is not removed. Therefore, representations of source and target point clouds in the space have different positions. Our registration is to remove the influence which can be regarded as a practical simulation of KSS mapping. When the two point clouds can be aligned into a same location in pre-shape space, it means that they can be mapped into a same reflection in KSS.}
  \label{f2}
\end{figure*}

To achieve the accurate registration result, the aforementioned influences should be reduced.
As a well-known method, Iterative Closest Point (ICP)~\cite{besl1992method} is proposed to reach the target. The ICP is a point-to-point scheme that searches the correspondence between point clouds by distance measurement. It is simple to implement and can be used in raw point clouds directly. However, the ICP and its variants ~\cite{fitzgibbon2003robust}\cite{granger2002multi} are likely to trap into the local optimum that reduces the accuracy of the registration process. Recently, deep learning frameworks ~\cite{aoki2019pointnetlk}\cite{sarode2019pcrnet} are used to build the registration methods. Such methods learn the correspondence of point clouds based on large database training with fast computation speed. However, the influences of similarity transformations can not be removed theoretically. The frameworks depend on species of training set. It cannot be guaranteed that
local or global features of input point cloud are learned from certain training set. Therefore, the classical deep learning frameworks have good performance in specific data set, but not in practice. For all methods mentioned before, different scales cannot be unified with an effective solution.

In this paper, we propose a new point cloud registration method, KSS-ICP, to achieve more accurate and robust registration results. It is inspired by Kendall shape space (KSS) theory~\cite{kendall1984shape}. The KSS is a manifold space, which is constructed by discrete point sequences. In KSS, influences of translation, scale, and rotation are removed. In other word, the KSS-based representation of point cloud is invariant to the similarity transformation which covers the mentioned influences. Benefited from such property, we propose a point cloud representation to map a point cloud into the KSS. The representation keeps the geometric consistency with the original point cloud while reducing influences of non-uniform density, locations, and scales. Based on the representation, we design an alignment process to achieve the best rotation parameters between point clouds. Combining the alignment and the ICP method, accurate registration can be achieved which is robust to similarity transformations, non-uniform density, noise, and defective parts. Some instances are shown in Figure \ref{f1} and \ref{f1_2}.  The contributions of our work are summarized as follows.

\begin{itemize}

\item A point cloud representation based on KSS is proposed. It provides a regular form for different point clouds, which can maintain a certain point number and achieves uniform point density. It therefore reduces influences of point density, locations, and scales.

\item An alignment for point clouds without defective or missing parts is designed based on the proposed point cloud representation. Benefited from the invariant property for similarity transformations, the registration result can be computed with a simple implementation in KSS. It is also robust to noise in point clouds without complicated feature extraction.

\item A partial-complete alignment scheme is constructed as a supplementary scheme to deal with incomplete geometric structures. With an additional searching in a candidate set, a point cloud with incomplete geometric structure can be aligned to the complete one scanned from the same 3D object.

\end{itemize}

The pipeline of KSS-ICP is shown in Figure \ref{f2}. The rest of the paper is organized as follows. In Sec. 2, we review existing classical methods for point cloud registration. In Sec. 3, we introduce the point cloud representation based on KSS, followed by the alignment process in Sec. 4. We demonstrate the effectiveness and efficiency of our method with extensive experimental evidence in Sec. 5, and Sec. 6 concludes the paper.

\section{Related Works}

There are a large number of papers for registration, which bring some difficulties to show the complete related works. We focus on the rigid point cloud registration task and introduce some representative methods in this part. Such methods are classified into three categories: distance metric-based registration, feature-based registration, and deep learning-based registration.

Distance metric-based registration methods achieve registration results by point distance optimization. The ICP and its variants belong to this category, including original ICP~\cite{besl1992method}, LM-ICP~\cite{fitzgibbon2003robust}, EM-ICP~\cite{granger2002multi}, Scale-ICP~\cite{ying2009scale}, GMM-ICP~\cite{jian2005robust}, BiK-ICP~\cite{yang2019point}, and Fast-ICP~\cite{Zhang2022FastICP}. More ICP variants are discussed in reviews~\cite{makela2002review}\cite{rusinkiewicz2001efficient}. The drawback of such methods is that the registration process traps into the local optimum with high probability. Some methods attempt to solve the problem based on Branch-and-Bound(BnB) scheme~\cite{olsson2008branch}. The BnB scheme is a tree structure-based algorithm and this is used to simplify alignment in $SO(3)$ (3D rotation group). The representative methods include $L2$ error optimization~\cite{li20073d}, stereographic projection~\cite{parra2014fast}, consensus set maximization~\cite{bazin2012globally}, camera pose alignment~\cite{enqvist2008robust}, and globally optimal solution(Go-ICP)~\cite{yang2013go}. However, most of them are sensitive to point clouds with different scales. With some large rotations($>45^\circ$ by each axis), the accuracy of registration may be reduced.

Feature-based registration methods build the correspondence of point clouds based on point features. Such methods avoid redundant searching in $SO(3)$ by feature alignment. In theory, the feature alignment keeps better geometric consistency during the registration process. The representative methods include normal distributions transform~\cite{biber2003normal}\cite{serafin2015nicp}, shape context~\cite{belongie2002shape}\cite{frome2004recognizing}, sub-maps~\cite{guan2009registration}, Rotational Projection Statistics features~\cite{guo2014accurate}, covariance matrices~\cite{tabia2015covariance}, and
point feature histograms~\cite{rusu2008aligning}\cite{rusu2009fast}\cite{yang2016fast}. The drawback of such methods is that the performance of registration is sensitive to the quality of the feature. Noise and defective parts in point clouds reduce the accuracy and robustness of features inevitably. For point clouds with large volumes, the huge calculation of feature extraction also affects the practicality of the methods.

Deep learning-based registration methods are becoming more popular recently. Using the deep correspondence from the data training, registration results can be achieved with the balance of efficiency and robustness. Such methods include PointNetLK~\cite{aoki2019pointnetlk}, Deep ICP~\cite{lu2019deepicp}, Deep Closest Point~\cite{wang2019deep}, PRNet~\cite{wang2019prnet}, IDAM~\cite{li2019iterative}, RPM-Net~\cite{yew2020rpm}, 3DRegNet~\cite{pais20203dregnet}, DGR~\cite{choy2020deep}, and PCRNet~\cite{sarode2019pcrnet}. Although there are so many works based on deep learning frameworks, some defects still exist in practice. Firstly, the deep learning framework trains and learns the deep correspondence between points from local patches matching in most cases. It tends to build a local but not global correspondence result, which reduces the performance in registration for point clouds with large difference in poses or rotations. Secondly, without reasonable pre-processing, the deep learning framework can not reduce the influence of different scales of point clouds. Finally, the deep learning framework is sensitive to the non-uniform density and this affects the accuracy of registration.

The KSS-ICP belongs to the first category. It searches the registration result in KSS. The influences of similarity transformations are reduced by KSS-ICP. With the proposed alignment in global view, KSS-ICP avoids the local optimum as much as possible. In the following sections, we discuss the details of KSS-ICP.

\section{Point Cloud Representation}

As mentioned before, the key property of KSS is that the influence of similarity transformation can be removed. For point cloud registration, the property can be used to reduce the influence of similarity transformation. The KSS provides a shape analysis tool to measure different shape models such as point clouds, images, and meshes. Once the point cloud representation is provided in KSS, the related shape analysis tool can be used to process the registration task without the influence of similarity transformation. In this section, we introduce the KSS theory and the necessary conditions of the representation in KSS. Following the conditions, we provide the point cloud representation
based on KSS, which is inspired by~\cite{lv20203d}.

\begin{figure}
  \centering
  \includegraphics[width=\linewidth]{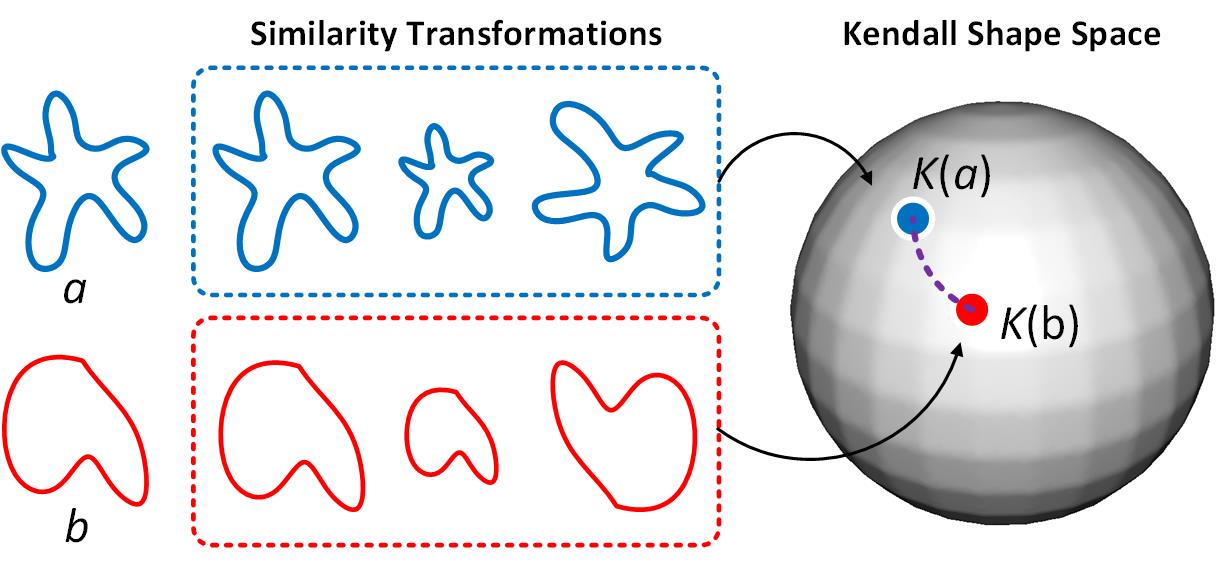}
  \caption{An instance of shape feature-based measurement in KSS. The 2D point sequences $a$ and $b$ takes different shape features. The shape feature should not be affected by the similarity transformations, which means that the objects in blue dotted box (or red one) have same shape feature. In KSS, such objects share same representation. Therefore, the geodesic distance (purple dotted curve) in KSS can be regarded as a reasonable shape feature-based measurement between $a$ and $b$.}
  \label{f_k}
\end{figure}

\subsection{Kendall Shape Space (KSS)}

Firstly, we provide a brief introduction for KSS. The KSS is a Riemannian manifold with the quotient space property. It can be used to represent the shape features of discrete point sequences. In KSS, a point sequence is invariant to similarity transformation. The mathematical definition of KSS is shown as
\begin{equation}
\textcolor{black}{
\begin{array}{c}
\textbf{M} = {R^{m \times k}}\backslash\left\{0\right\},\\
\textbf{M}/G = \textbf{K},
\end{array}}
\end{equation}
where $\textbf{M}$ is a Riemannian manifold constructed by discrete point sequences. The dimensions of $\textbf{M}$ are $m \times k$ ($k$ points with $m$ dimensions). The $G$ is a group of similarity transformation. $\textbf{K}$ represents the KSS constructed by $\textbf{M}/G$, which is still a Riemannian manifold. The operator $/$ is the quotient group computing. It means that $\textbf{M}$ is mapped into $\textbf{K}$ while removing the influence of $G$. Once we define and implement the mapping, the influence of similarity transformation is removed. An instance is shown in Figure~\ref{f_k}. To map a discrete point sequence into the KSS, a set of pre-processing operations are needed:
\begin{equation}
\begin{array}{c}
{K_{\rm{s}}}(a) = ({x_{(1)}} - \bar x,...,{x_{(k)}} - \bar x)/s(a),\\
\bar x = \frac{1}{k}\sum\limits_{j = 1}^k {{x_{(j)}}} ,s(a) = {(\sum\limits_{j = 1}^k {\left\| {{x_{(j)}} - \bar x} \right\|} )^{1/2}},
\end{array}
\label{e2}
\end{equation}
where $a$ is a discrete point sequence with point number $k$. Using the center $\bar{x}$ to uniform the discrete point sequence, the influence of different locations and scales is removed. A middle shape space $\textbf{K}_\textbf{s}$ between $\textbf{M}$ and $\textbf{K}$ is achieved, called pre-shape (i.e., not KSS yet) space~\cite{kendall1984shape}. In most cases, it is difficult to achieve the representation of discrete point sequence in $\textbf{K}$. However, the representation and related measurement in $\textbf{K}$ can be simulated in $\textbf{K}_\textbf{s}$ (as a sufficient approximation in practice), represented as
\begin{equation}
{d_{\textbf{K}}}\left( {K(a),K(b)} \right) = \mathop {\inf }\limits_{O \in SO(3)} {d_{{\textbf K}_{\textbf s}}}\left({{K_s}(a),O\cdot{K_s}(b)} \right),
\label{e3}
\end{equation}
where $d_{K}$ is the measurement in $\textbf{K}$, $d_{K_s}$ is the measurement in $\textbf{K}_\textbf{s}$, and $b$ is another discrete point sequence. $K_s(a)$ and $K_s(b)$ are representations of $a$ and $b$ in $\textbf{K}_\textbf{s}$. $O$ is the rotation from $SO(3)$, which is used to find a match between $K_s(a)$ and $K_s(b)$.
The operator $\cdot$ means that the former transformation is implemented into the latter object. In this place, it means to rotate $K_s(b)$ by $O$. In most cases, it is difficult to achieve representations ($K(a)$ and $K(b)$) of discrete point sequences and related measurement ($d_{\textbf{K}}$) in $\textbf{K}$. Equation \ref{e3} provides a roundabout solution to compute the measurement in $\textbf{K}_\textbf{s}$. Once the representation of discrete point sequence in $\textbf{K}_\textbf{s}$ is provided, the computation of $d_{\textbf{K}}$ can be transferred to find the rotation $O$ and this defines a transformation matrix between two discrete point sequences. The transformation matrix can be used to achieve the registration result.

According to the description of KSS theory, there are some requirements for construction of representation in $\textbf{K}_\textbf{s}$: 1. point numbers should be equal; 2. weights of points should be equal; 3. the center should be aligned. In Equation \ref{e2}, discrete point sequences should have same point number $k$ to achieve the representation in $\textbf{K}_\textbf{s}$. Each point in the discrete point sequence has same weight. It means that the point cloud should have uniform density. The center should be aligned, otherwise the representations in $\textbf{K}_\textbf{s}$ are not accurate. In order to meet the above requirements, we provide the construction of point cloud representation in $\textbf{K}_\textbf{s}$.

\begin{figure}
  \includegraphics[width=\linewidth]{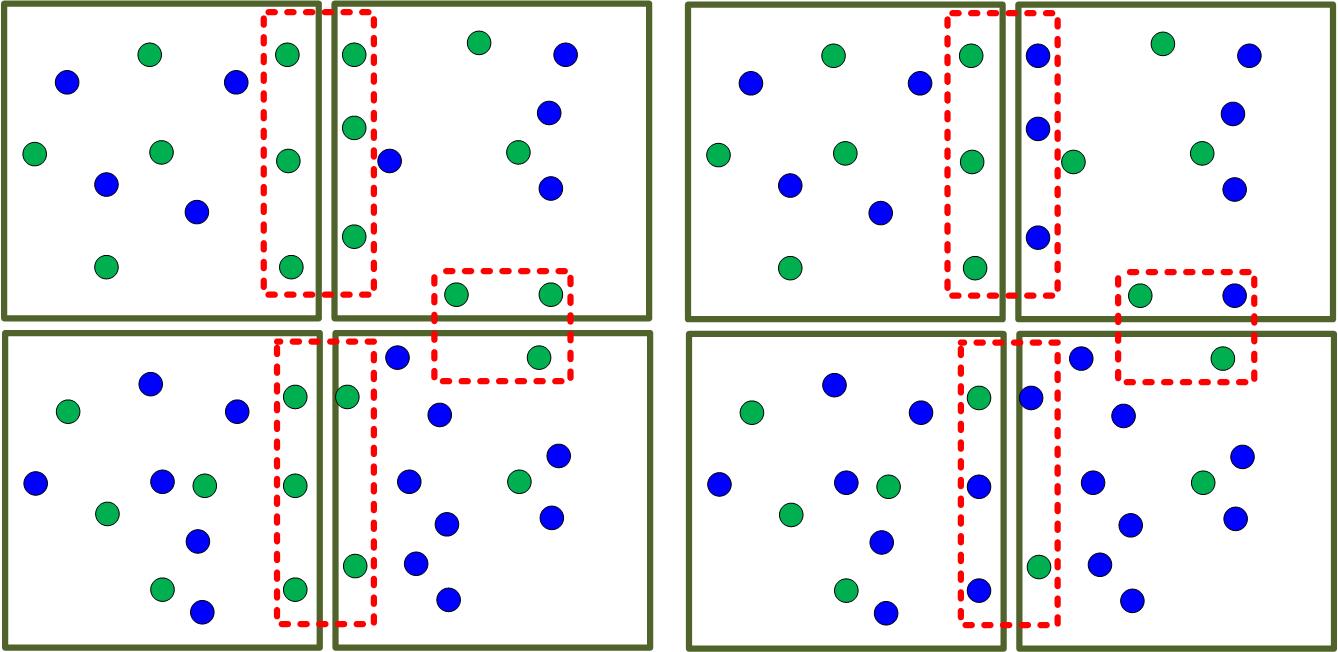}
  \caption{An instance of non-uniform density in a point cloud. Blue points are original points. Green points are resampling results. Left: points with non-uniform density around the boundary (red dotted box); Right: non-uniform density is removed by parallel resampling.}
  \label{f3}
\end{figure}

\subsection{Representation in $\textbf{K}_\textbf{s}$}

\begin{figure}
  \includegraphics[width=\linewidth]{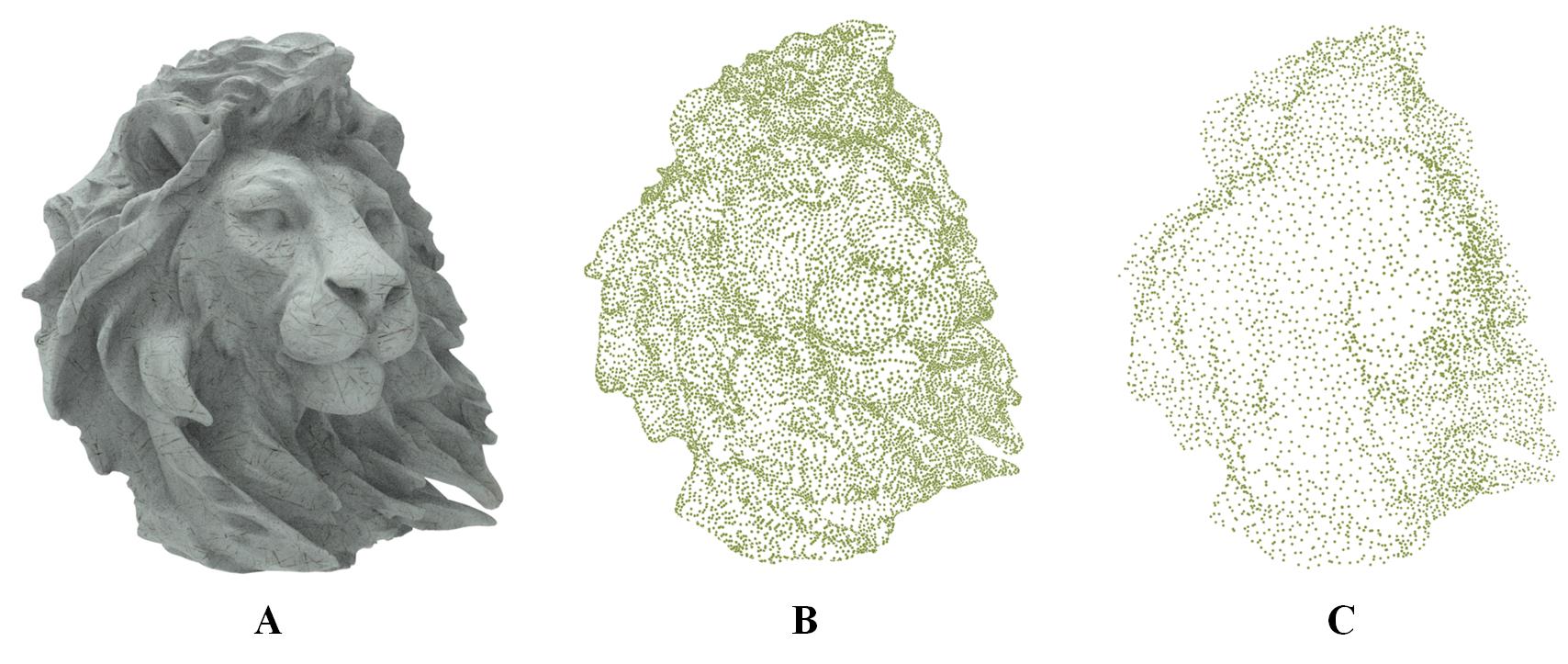}
  \caption{An instance of simplified point clouds. A: real 3D object; B: scanning point cloud; C: simplified point cloud (2,000 points).}
  \label{f4}
\end{figure}

The construction of the point cloud representation in $\textbf{K}_\textbf{s}$ includes two parts: 1. simplify input point cloud into an alternative one with same point number and uniform density; 2. map the alternative point cloud into $\textbf{K}_\textbf{s}$ as the point cloud representation. For the first part, we utilize a parallel simplification~\cite{Lv2021Approximate} to achieve the alternative point cloud. Basically, the simplification can be regarded as a parallel version of Farthest point sampling (FPS). It is used to adjust point cloud density and control point number efficiently. It includes three steps. Firstly, we split the input point cloud into a voxel structure and this is constructed by voxel boxes with the same scale. Then, we use the local FPS to parallel simplify the point cloud in different voxel boxes. Finally, we combine the simplification results from different voxel boxes to achieve the
simplified point cloud. The simplify point number $|P_{v}'|$ in voxel box $v$ is provided as
\begin{equation}
|P_{v}'| = |P_v| \times k / |P|,
\label{e5}
\end{equation}
where $|P|$ is the point number of point cloud $P$, $|P_v|$ is the point number in $v$. The scale of the voxel box is computed as
\begin{equation}
\label{e6}
V_s = L_p/\left[\sqrt[3]{{|P|}}/2\right],
\end{equation}
where $L_p$ is the longest edge's length of the bounding box from $P$. Equation \ref{e6} is achieved based on previous practical experience. It ensures that there are enough points in different voxel boxes for parallel simplification. Besides, the parallel simplification should not be processed in adjacent voxel boxes at the same time. The reason is that such a situation produces non-uniform density in the boundary. An instance is shown in Figure \ref{f3}. To avoid the situation, we set different rounds for parallel simplification. In the same round, the voxel boxes used for simplification have not adjacency to each other. During the simplification, the method considers points have been simplified in adjacent boxes to keep the uniform density in the boundary. The simplification result can be regarded as the simplified point cloud with certain point number, uniform density, and accurate geometric consistency. An instance is shown in Figure \ref{f4}. The details of the implementation can be found in~\cite{Lv2021Approximate}.

\begin{figure}
  \centering
  \includegraphics[scale=0.5]{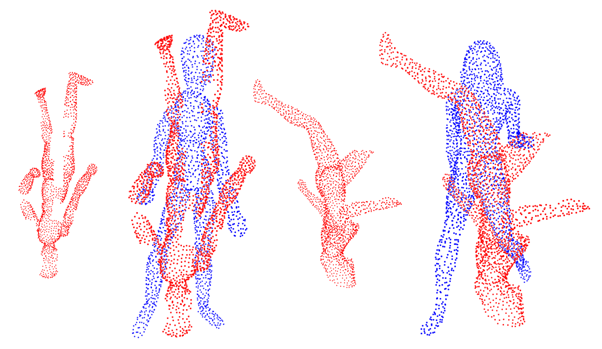}
  \caption{Instances of point cloud representation. Except for rotation, influence of point density, translation and scaling is removed from the representation.}
  \label{f5}
\end{figure}

Based on the simplified point cloud, we provide the second part to achieve the representation in $\textbf{K}_\textbf{s}$. With same point number and uniform density, the simplified point cloud can be used as the input data in Equation \ref{e2}. Then, we achieve the point cloud representation in $\textbf{K}_\textbf{s}$. The influences of different locations and scales are reduced. In Figure \ref{f5}, two instances of point cloud representation are shown. To achieve the final correspondence result, we provide an alignment based on the point cloud representation. The details are discussed in the following part.

\begin{figure}
  \centering
  \includegraphics[width=\linewidth]{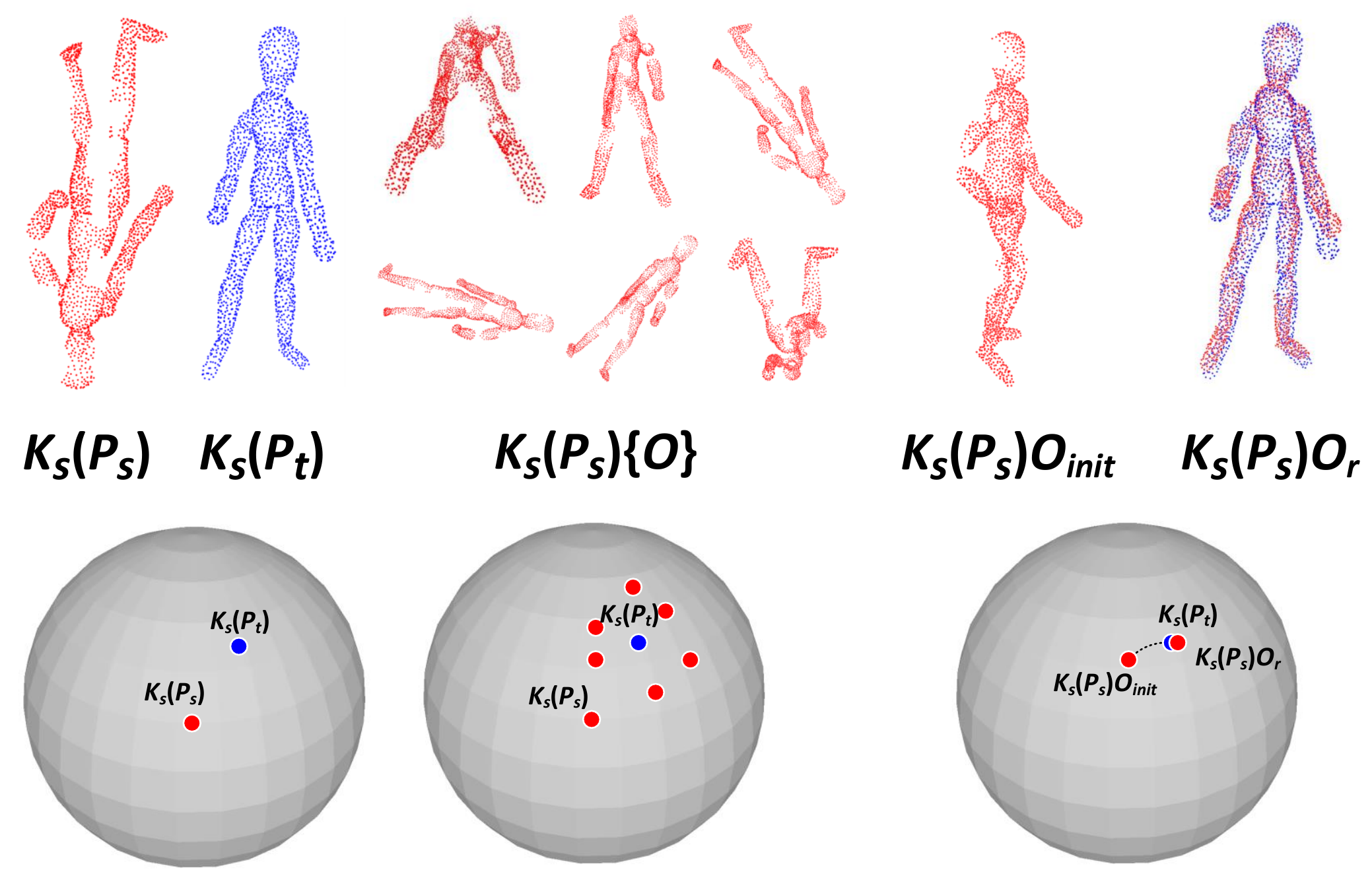}
  \caption{An instance of alignment based on $SO(3)$ searching in $\{O\}$.}
  \label{f6}
\end{figure}

\section{Alignment}

\subsection{Global-Global Alignment}
To achieve the registration result, an alignment process should be provided to search the rotation $O$ in Equation \ref{e3}, which is based on the point cloud representation. Once the rotation is achieved, it can be regarded as the transformation matrix for registration. In~\cite{kendall1984shape}, an alignment method was provided, called Procrustes analysis. In Equation \ref{e7}, the Procrustes analysis is shown as
\begin{equation}
\label{e7}
\begin{array}{c}d_{\textbf{K}}(K(a),K(b))=arc\cos(tr\Lambda),\\
K_s(a)K_s(b)^t=U\Lambda V,\\
\end{array}
\end{equation}
where the distance $d_{K}$ between $K(a)$ and $K(b)$ in $K$ is computed by their vector-based angle. The value of the angle equals to the arc cosine of matrix trace $tr\Lambda$, $\Lambda$ is computed from the singular value decomposition (SVD) of the $K_s(a)K_s(b)^t$ with a pair of unitary matrices $U$ and $V$. Essentially, the Procrustes analysis is a process to compute SVD for measurement between KSS-based measurement. Using the SVD result, the rotation can be reconstructed. It is represented as
\begin{equation}
\label{e8}
O_r=(UV)^t,
\end{equation}
where $O_r$ is the rotation which can be regarded as the transformation matrix. Unfortunately, it cannot be used in point cloud registration directly. The reason is that the Procrustes analysis requests the point cloud representation should be ordered. In most cases, the requirement can not be satisfied. To solve the problem, we propose a partitioned $SO(3)$ searching scheme to be a global-global alignment for registration. It is similar to $SO(3)$ searching by BnB scheme~\cite{yang2013go}. The alignment energy is formulated as
\begin{equation}
\label{e9}
E_d=H(K_s(a),O\cdot K_s(b)),
\end{equation}
where $H$ is the Hausdorff distance, and other parameters have been introduced in Equation \ref{e3}. As a classical measurement, the Hausdorff distance can be used to provide a quantitative analysis for similarity between two 3D point clouds~\cite{taha2015efficient}. It does not require ordered point clouds as input and can provide reasonable measurement in global view. To achieve the optimization result, we build a candidate set $\{O\}$ and this is achieved from the rotation with certain angle $\theta$. Selecting the $O_i$ with minimum value of $E_d$ to be the initial rotation $O_{init}$. The alignment process based on $\{O\}$ is shown as
\begin{equation}
\label{e10}
\begin{array}{c}
O_{init}=\{O_i\vert minE_d(O_i)\},O_i\in\{O\},\\{\{O\}}=\{(\theta.x,\theta.y,\theta.z)\vert x,y,z\in\lbrack0,2\mathrm\pi/\mathrm\theta\rbrack\},
\end{array}
\end{equation}
where $\theta.x$ means to rotate around the $X$ axis with $\theta.x$ degrees, $x$ is an integer, $x \in \lbrack0,2\mathrm\pi/\mathrm\theta\rbrack$. $\theta.y$ and $\theta.z$ share the similar definition to $\theta.x$. According to different rotations around the axes of $X$, $Y$, and $Z$, we achieve the candidate set $\{O\}$. Then the $O_{init}$ can be obtained. By default, $\theta = \pi/6$ and the $\{O\}$ includes $1728 (12\times12\times12)$ items. Based on the $O_{init}$, we use ICP to perform precise alignment. Then our alignment with a coarse to fine process is represented as
\begin{equation}
\label{e11}
O_r=T\cdot O_{init},
\end{equation}
where $T$ is the transformation matrix computed by ICP. Combining $T$ and $O_{init}$, the final rotation $O_r$ for registration is computed, which can be regarded as the registration result. In Figure \ref{f6}, an instance of registration result is shown by our alignment. It can be regarded as the 3D version of the instance shown in Figure~\ref{f_k}.

\begin{figure}
  \includegraphics[width=\linewidth]{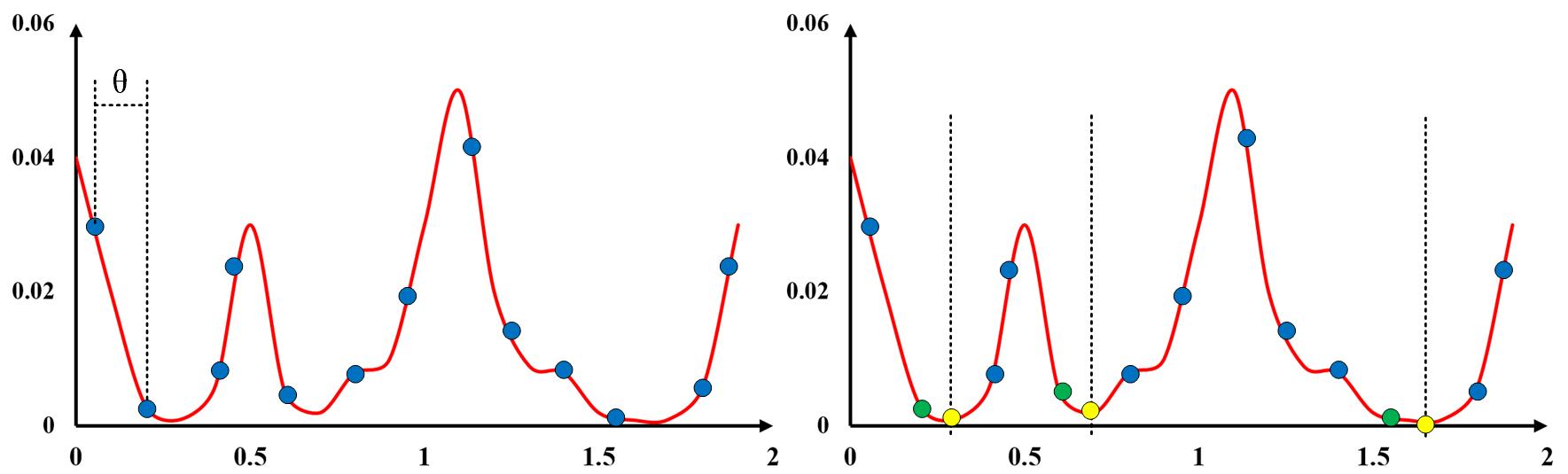}
  \caption{Energy curve changes of rotation searching by additional process. Vertical axis: $E_d$; Horizontal axis: changing rotations by $\theta$. The blue points represent the rotation in $\{O\}$. The green points represent the local minimum rotations from local optimal set $\{O_{local}\}$. The yellow points represent the accuracy local minimum rotations by ICP. }
  \label{f62}
\end{figure}

\begin{figure}
  \includegraphics[width=\linewidth]{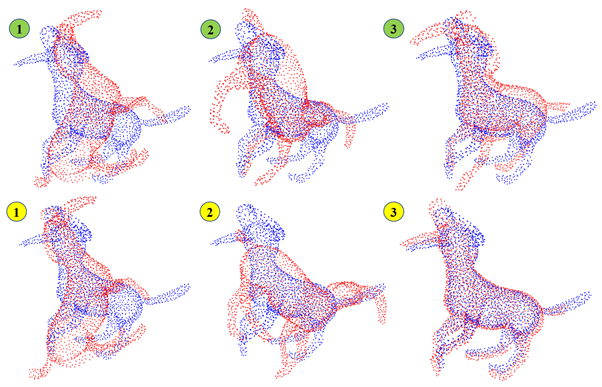}
  \caption{An instance of additional process for rotation searching is shown. First row: the preliminary registration results from local minimum set, which correspond to the green points in Figure \ref{f62}; Second row: the accuracy local minimum registration results by ICP, which correspond to the yellow points in Figure \ref{f62}.}
  \label{f63}
\end{figure}

Based on the point cloud representation, the alignment improves the convergence performance of ICP and achieves more accurate registration result. However, it should be noticed that the alignment is not a strictly global optimization for $SO(3)$ searching. The candidate set $\{O\}$ just includes some discrete rotations. Although it is effective for most cases in practice, some error rotation results can not be avoided, especially for point clouds with symmetrical structures. To solve the problem, we add an additional process if the $E_d$ is larger than a threshold $Q_{the}$ (by default, $Q_{the}=0.001$). The additional process is to search other local optimal results. The global optimal result is achieved from the local optimal set. The additional process is formulated as
\begin{equation}
\label{e12}
\begin{array}{c}
E_d(T.O_{init})>Q_{the}, O_r=\{O\vert min\{E_d(T\cdot O_j)\}\},\\
O_j\in\{O_{local}\},
\end{array}
\end{equation}
where $\{O_{local}\}$ is the local optimal set, which is constructed by the rotation $O_j$. We define a searching kernel in $\{O\}$ to represent the range. $O_j$ belongs to $\{O_{local}\}$, which is the local optimal result with the minimum $E_d$ in a searching kernel. By default, the scale of kernel is 125 ($5\times5\times5$). Using ICP on the local optimal set, the final rotation $O_r$ is achieved. In Figure \ref{f62}, we show the $E_d$ changing in additional process. The green points represent values of $E_d$ from $\{O_{local}\}$. The yellow points are results processed by ICP on green points. The $O_r$ is achieved from the $\{O_{local}\}$ with minimum $E_d$. In order to explain the process more intuitively, we show an instance of local registration results and corresponding accuracy results by ICP in Figure \ref{f63}.

\begin{algorithm}[t]
  \caption{Implementation of KSS-ICP (Global-Global)}
  \label{A1}
  \begin{algorithmic}[1]
    \Require
    Input source $P_s$ and target $P_t$ point clouds.
    \State Voxelization for $P_s$ and $P_t$ with the scale $V_s$ (Equ.\ref{e6}).
    \State Distribute voxel boxes into different rounds.
    \State Set $k$ (Equ.\ref{e5}).
    \State Initial adaptive point clouds $P_{sims}$ and $P_{simt}$:
    \For i rounds
    \State Compute resampling point number $|P_{v}'|$ (Equ.\ref{e5}).
    \State Add resampling points from previous rounds.
    \State Parallel resample Voxel boxes by FPS.
    \State Add resampling points into $P_{sims}$ and $P_{simt}$.
    \EndFor
    \State Output $P_{sims}$ and $P_{simt}$ into (Equ.\ref{e2}).
    \State Achieve representations $K_s(P_{s})$ and $K_s(P_{t})$.
    \State Rotate ($\theta.x,\theta.y,\theta.z$) to build $\{O\}$.
    \For i rotations $\{O\}$
    \State rotate $K_s(P_{s})$ by $O_i$ .
    \State Compute $E_d$ from $O_i.K_s(P_{s})$ and $K_s(P_{t})$ (Equ.\ref{e9}).
    \EndFor
    \State Output $O_{init}$ with minimum $E_d$.
    \State Use ICP for $O_{init}$ (Equ.\ref{e11}), output $O_r$.
    \State Additional process to update $O_r$ if needed.
    \Ensure Output $O_r$.
  \end{algorithmic}
\end{algorithm}

Combining the point cloud representation and alignment, the KSS-ICP can be implemented for point clouds with complete geometric structure. Some  details of the implementation are shown in Algorithm \ref{A1}. It is robust to similarity transformations, including different locations, scales, and rotations. The alignment with additional process does not require complex optimization to achieve the best rotation. Without complex point feature analysis and data training, It achieves accurate registration results in practice.

\subsection{Partial-Global Alignment}
\begin{figure}
  \includegraphics[width=\linewidth]{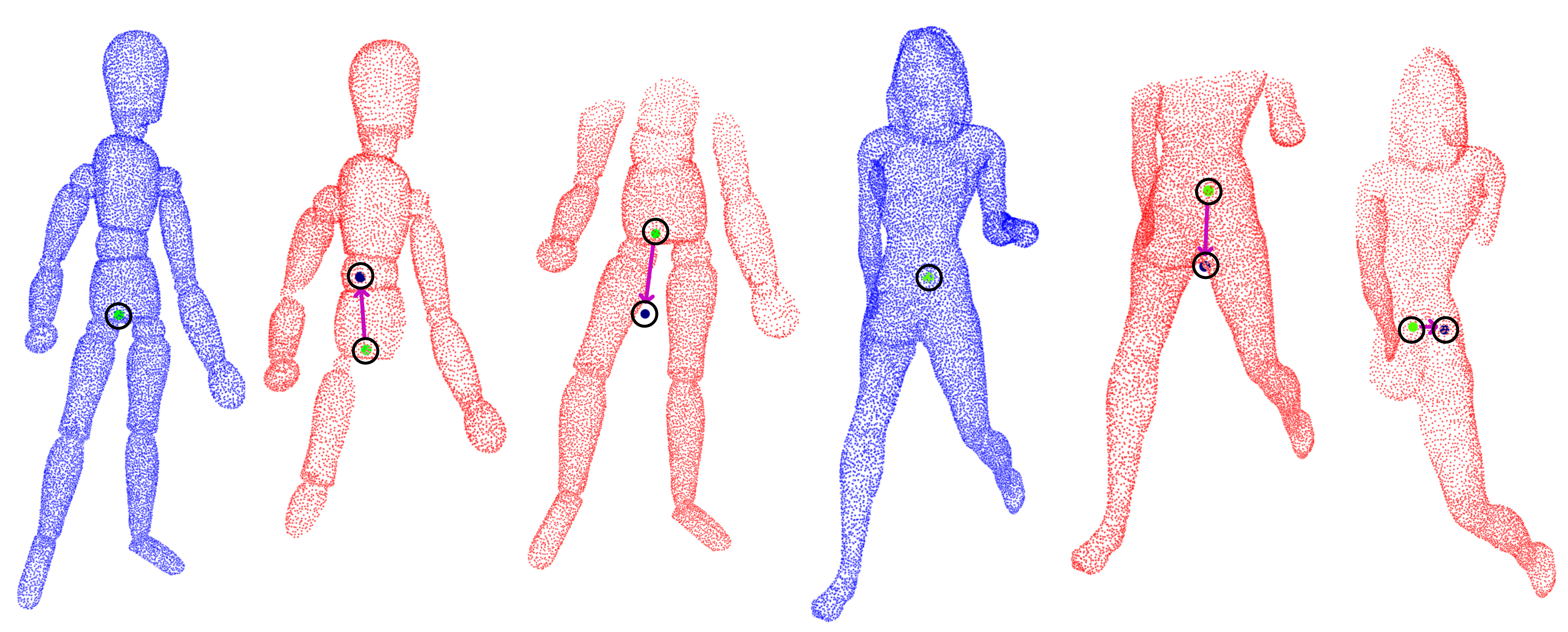}
  \caption{Instances of point clouds with global geometric structures (blue) and partial geometric structures (red). The arrows represent displacements between centers of point clouds with global and partial geometric structures.}
  \label{f64}
\end{figure}

The global-global alignment can be used to align point clouds with complete geometric structures. The center extracted from the point cloud is stable and reliable, even the input point cloud has non-uniform density. In our previous work~\cite{Lv2022Intrinsic}, the performance of the alignment with a rough implementation has been verified. However, centers of point clouds with incomplete geometric structures cannot be aligned. Therefore, the performance of the alignment is reduced for point clouds with incomplete geometric structures or defective parts. Unfortunately, a large number of point clouds are scanned from single view of the real 3D objects or scenes. Incomplete geometric structures are produced in such cases. In Figure \ref{f64}, some instances are shown. To solve the problem, we propose a partial-global alignment to be a refinement for our alignment.

\begin{figure}
  \centering
  \includegraphics[scale=0.4]{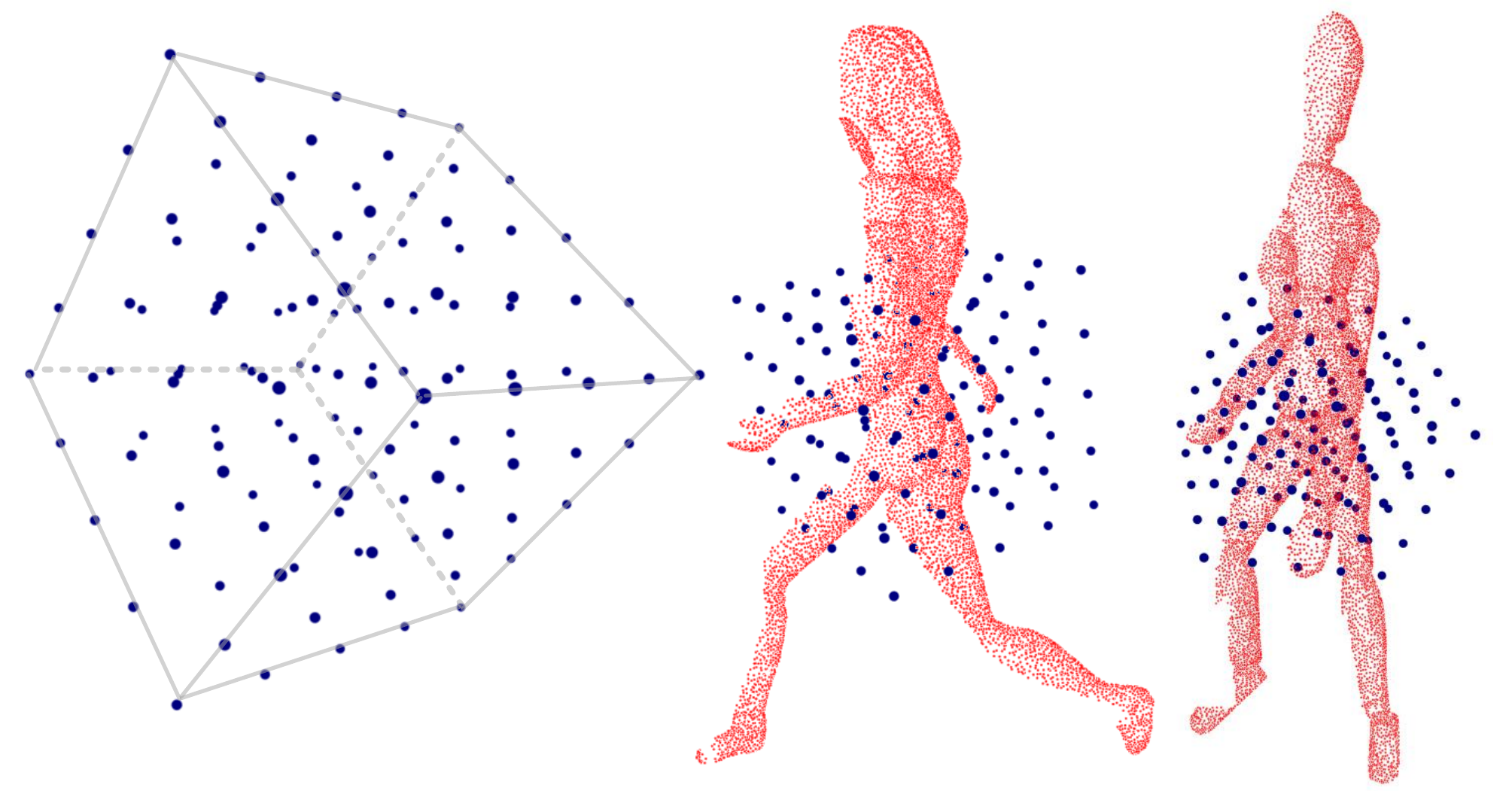}
  \caption{Instances of center searching regions in point clouds with incomplete geometric structure. Blue points are selected in the searching region (grey box) to construct candidate center set.}
  \label{f65}
\end{figure}

The basic idea of our partial-global alignment is to expand searching regions in KSS with different potential centers. For example, there have a pair of point clouds scanned from a 3D object. One point cloud has some missing parts, and the other one has the complete geometric structure. An unknowable displacement exists between centers of the two point clouds. The center displacements are shown in Figure \ref{f64}. To solve the problem, we build a searching region which includes a set of candidate centers. In the region, we select a center from the set to instead the original one in point cloud with incomplete geometric structure. The new center has the optimal matching result for registration. Then, the influence of the displacement can be reduced. In Figure \ref{f65}, we show some instances of searching region and candidate center set.

We provide the details of implementation for partial-global alignment. Let ${C}$ be the set of candidate centers and $c_i$ is a center in ${C}$. The alignment energy discussed in Equation \ref{e9} is changed as
\begin{equation}
\label{e122}
E_{cd}=H(K_s(a),O\cdot K_s(c_i.b)),c_i\in \{C\},
\end{equation}
where $a$ and $b$ are two point clouds with complete and incomplete geometric structures, $c_i.b$ means that the center of $b$ is replaced by $c_i$. We set a square box to be the searching region and sample different candidate centers according to fixed step. In our practice, we construct a local coordinate system to define the square box. The construction steps of local coordinate system includes: 1. compute the current center $c_r$; 2. search the point $p_y$ with maximum distance to $c_r$, obtain unit vector of $p_yc_r$ as local $Y$ axis; 3. search the point $p_x$ which has minimum distance to $c_r$ and hasn't linear relationship with $p_yc_r$, obtain vector $p_xc_r$, and use unit vector of $p_xc_r\times p_yc_r$ as local $Z$ axis; 4. achieve local $X$ axis by $Y\times Z$. Based on the local coordinate system, the candidate centers can be collected:
\begin{equation}
\label{e123}
\{C\} = \{c_r+(n_xX,n_yY,n_zZ)\},
\end{equation}
where $n_x$, $n_y$ and $n_z$ are movements based on local coordinate system. With different values of $n_x$, $n_y$ and $n_z$, candidate centers are generated into $\{C\}$. Values of $n_x$, $n_y$ and $n_z$ can be specified by user according to certain application. In our practice, we assume that the point cloud with incomplete geometric structure contains more than half of the global geometric feature. We set a range for $n_x$, $n_y$ and $n_z$, $n_x = n\cdot r_s/2, n=\{-2,-1,0,1,2\}$. $n_y$ and $n_z$ share the same range. $r_s$ is the radius of the searching region, $r_s=|p_yc_r|/4$. According to the sample scheme, we achieve 125 ($5\times 5\times 5$) candidate centers in $\{C\}$. It covers most cases of center's movement for the point cloud with incomplete geometric structure that produced by single-view scanning. Based on the ${C}$ and ${O}$, the initial partial-global alignment is provided as
\begin{equation}
\label{e13}
\begin{array}{c}
(c_{init}, O_{init})=\{(c_i, O_j)\vert minE_{cd}(c_i, O_j)\},\\
c_i\in\{C\}, O_j\in\{O\},
\end{array}
\end{equation}
where $c_{init}$ is the new center with optimal matching result for registration. Combining $c_{init}$ and $O_{init}$ (Equation \ref{e10}), we obtain an initial registration result. Similar to the global-global alignment, we use ICP and additional process to achieve more accurate result based on initial registration. Then, the final registration result can be achieved.

\subsection{Parallel Acceleration}

\begin{figure}
  \centering
  \includegraphics[width=\linewidth]{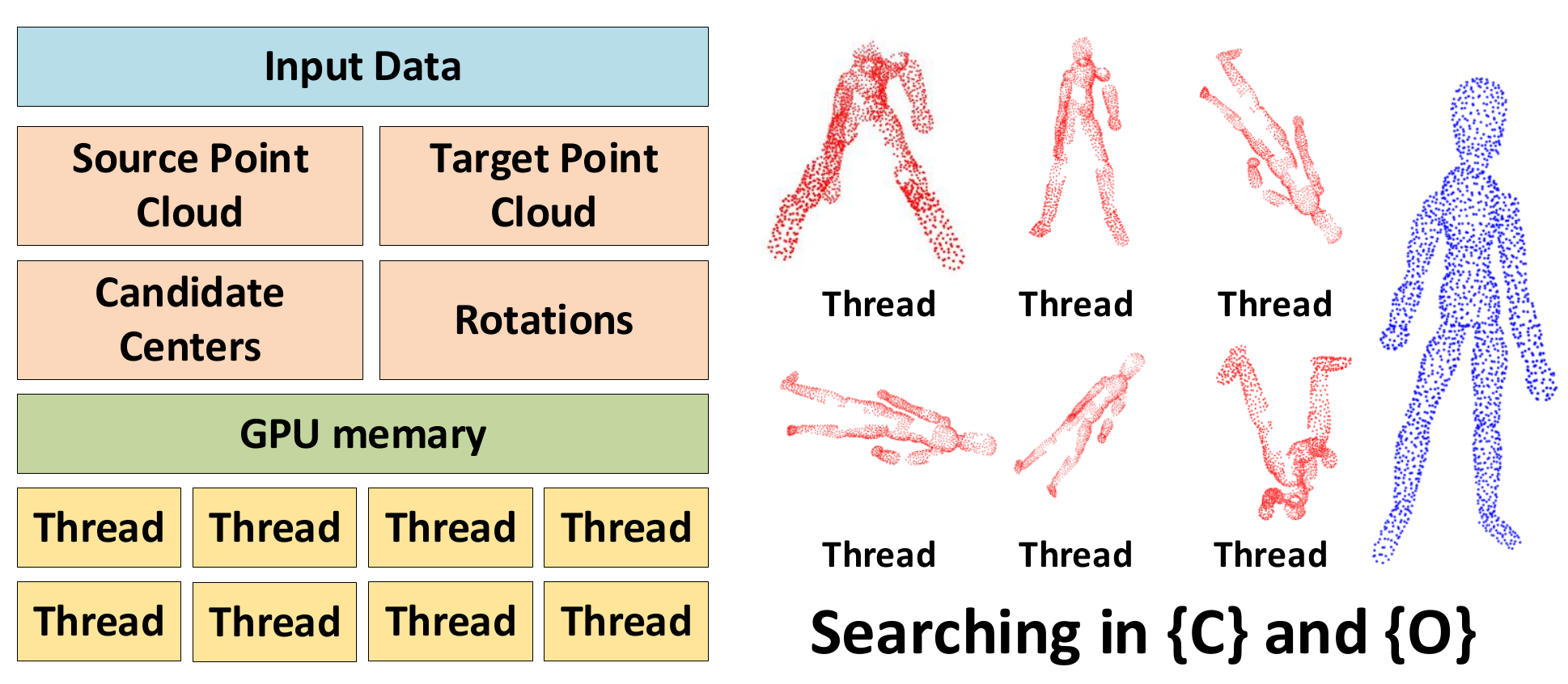}
  \caption{GPU-based parallel structure for alignment. $\{C\}$ is the candidate center set, $\{O\}$ is the candidate rotation set.}
  \label{f66}
\end{figure}

The alignment can be regarded as an initial registration before accurate registration by ICP. It is an exhaustive strategy based on a controlled searching region (defined by $\{C\}$ and $\{O\}$ in KSS). The quality of the alignment depends on the range and density of $\{C\}$ and $\{O\}$. Expanding the accuracy of $\{C\}$ and $\{O\}$ can improve the performance of initial registration. However, the time cost is huge even the initial registration has been controlled in KSS. Fortunately, searching initial registration result in $\{C\}$ and $\{O\}$ is a discrete matching process without dependency. It can be accelerated by parallel structure. In Figure \ref{f66}, an instance is shown to explain the alignment improved by the parallel structure, as further described next.

\begin{algorithm}[t]
  \caption{Alignment in a computing unit of GPU}
  \label{A2}
  \begin{algorithmic}[1]
    \Require
    Input $P_s$, $K_s(P_t)$, $\{C\}$ and $\{O\}$.
    \State Achieve current index of the unit.
    \State Computing the related $c_i$ and $O_j$ by the index.
    \State Computing $O_j.K_s(c_i.P_s)$.
    \State Computing $E_{cd}$ by Equation \ref{e122}.
    \Ensure Output $E_{cd}$.
  \end{algorithmic}
\end{algorithm}

We provide a GPU-based parallel acceleration to improve the efficiency of alignment. Each computing unit of GPU corresponds to a rotation $O_i$ of global-global alignment or a pair of parameters ($c_i,O_j$) of partial-global alignment. In GPU memory, the stored data just include point clouds which have been simplified, the candidate center set $\{C\}$, and the rotation set $\{O\}$. The time cost of data transfer from host memory to GPU memory can be ignored. In Algorithm \ref{A2}, we show the program of alignment in a computing unit. In most cases, the number of ${C}\times {O}$ ($125\times 1728$) is smaller than the sum of threads of a GPU with mainstream performance. It means that the initial alignment can be finished in one computation cycle of the parallel structure, which ensures the performance of our method. In experiment, we will show the improvement of the KSS-ICP on various datasets.

\begin{figure*}
  \centering
  \includegraphics[width=0.95\linewidth]{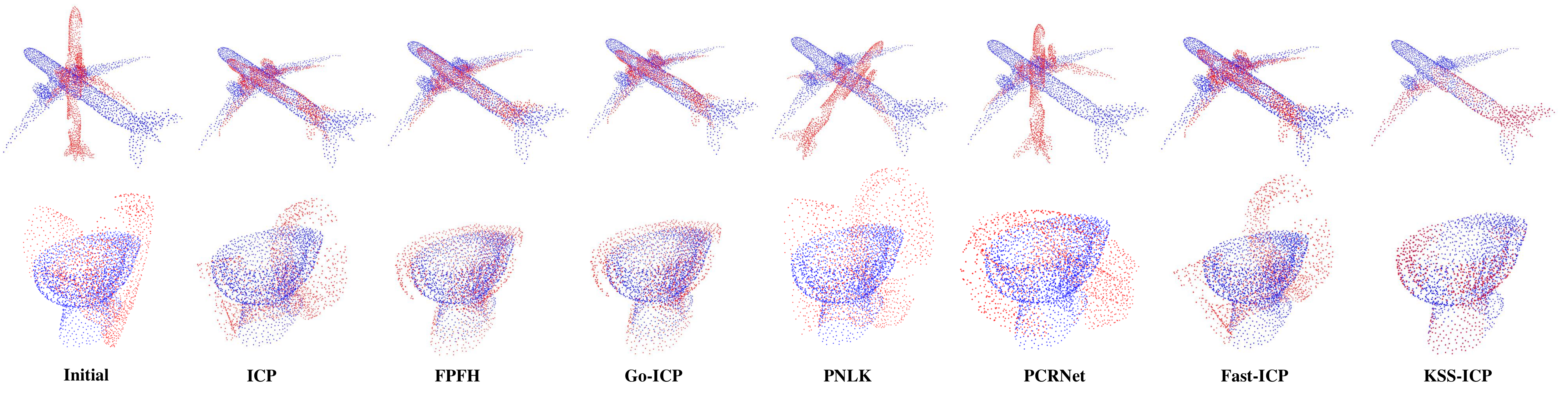}
  \caption{Comparisons of different registration results in ModelNet40 models.}
  \label{fe1}
\end{figure*}

\begin{figure*}
  \centering
  \includegraphics[width=0.95\linewidth]{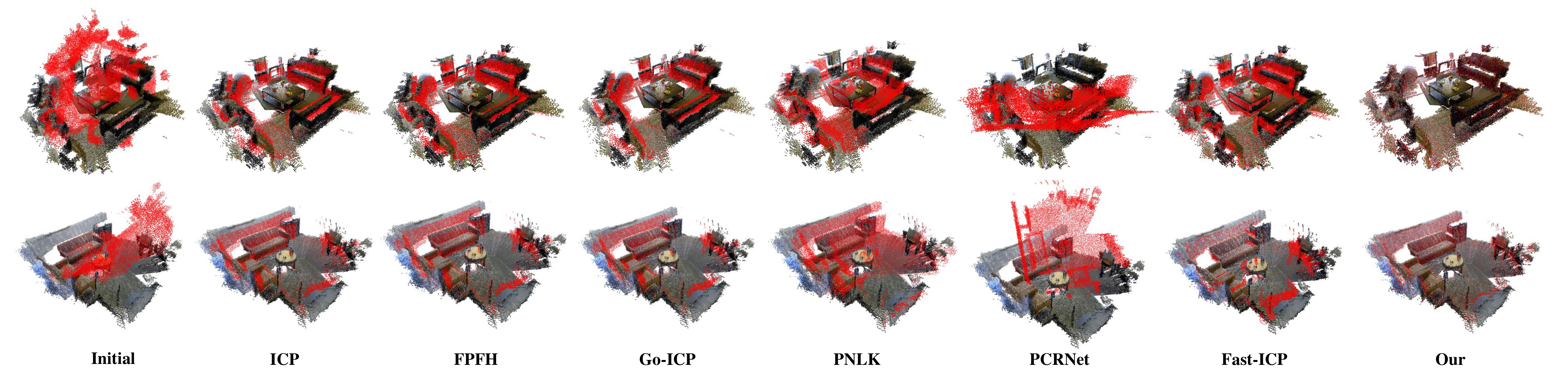}
  \caption{Comparisons of different registration results in RGB-D Scenes.}
  \label{fe2}
\end{figure*}

\begin{table}[t]
\caption{Evaluations for different registration methods in ModelNet40.}
\scriptsize
\label{ta1}
\begin{center}
\resizebox{\columnwidth}{!}{
\begin{tabular}{c|cccccc}
\toprule
\textbf{Method} & \textbf{MSE}      & \textbf{RMSE}     & \textbf{MAE}  & \textbf{MSE(R)} & \textbf{RMSE(R)} & \textbf{MAE(R)}      \\ \midrule
\textbf{ICP}             & 1.01E-2          & 8.36E-2          & 6.66E-2 & 6.22E-1     & 7.41E-1      & 6.05E-1        \\
\textbf{FPFH}            & 2.26E-3          & 3.94E-2          & 3.31E-2 & 3.12E-1     & 5.09E-1      & 3.53E-1         \\
\textbf{Go-ICP}          & 1.39E-3          & 3.22E-2          & 2.65E-2  & 2.85E-1    & 4.96E-1      & 3.34E-1       \\
\textbf{PNLK}           & 2.91E-2          & 1.44E-1          & 1.15E-1 & 8.14E-1     & 8.78E-1      & 7.41E-1          \\
\textbf{PCRNet}         & 3.05E-2          & 1.52E-1          & 1.22E-2 &8.14E-1     & 8.78E-1      & 7.66E-1       \\

\textbf{Fast-ICP}      & 1.19E-2&8.91E-2 &6.23E-2  & 5.69E-1     & 7.11E-1      & 5.63E-1 \\
\textbf{KSS-ICP}         & \textbf{3.89E-4} & \textbf{1.05E-2} & \textbf{8.44E-3}  & \textbf{1.01E-1}     & \textbf{1.73E-1}      & \textbf{1.21E-1} \\\bottomrule
\end{tabular}}
\end{center}
\end{table}

\begin{table}[t]
\caption{Evaluations for different registration methods in RGB-D Scenes dataset.}
\scriptsize
\label{ta2}
\begin{center}
\resizebox{\columnwidth}{!}{
\begin{tabular}{c|cccccc}
\toprule
\textbf{Method} & \textbf{MSE}      & \textbf{RMSE}     & \textbf{MAE} & \textbf{MSE(R)}      & \textbf{RMSE(R)}     & \textbf{MAE(R)}  \\ \midrule
\textbf{ICP}            & 1.45E-2           & 9.91E-2           & 7.27E-2 & 5.67E-1          & 7.18E-1         & 5.64E-1           \\
\textbf{FPFH}            & 1.62E-2          & 1.01E-1          & 7.49E-2& 5.29E-1          & 6.96E-1         & 5.42E-1             \\
\textbf{Go-ICP}          & 8.33E-3          & 7.42E-2           & 5.53E-2 & 4.57E-1          & 6.43E-1         & 4.91E-1            \\
\textbf{PNLK}           & 2.17E-1          & 3.59E-1           & 2.82E-1& 7.73E-1        & 8.41E-1         & 7.27E-1            \\
\textbf{PCRNet}         & 1.67E-1          & 3.85E-1          & 3.05E-1 & 8.31E-1          & 9.03E-1         & 8.03E-1  \\
\textbf{Fast-ICP}      & 5.81E-2& 1.72E-1 & 1.17E-1  & 5.17E-1     & 6.89E-1      & 5.37E-1 \\
KSS-ICP            &\textbf{4.47E-4} &\textbf{2.07E-2}  &\textbf{1.81E-2} &\textbf{3.94E-1} &\textbf{5.87E-1} &\textbf{4.49E-1} \\ \bottomrule
\end{tabular}}
\end{center}
\end{table}

\section{Experiments}

We implement the KSS-ICP on a machine equipped with Intel Xeon W2133 3.6GHz, 32GB RAM, Quadro P620, and with Windows 10 as its running system and Visual Studio 2019 (64 bit) as the development platform. The experimental point cloud models were selected from ModelNet40\cite{wu20153d}, RGB-D Scenes\cite{lai2014unsupervised}, and SHREC\cite{bronstein2010shrec} datasets. The test dataset from ModelNet40 contains 1235 models (since there are many similar models, we select models with 10\% sampling rate from each category); the test dataset from RGB-D scenes contains all scene models (14); the test dataset from SHREC contains all models (1200 models from 50 categories). Firstly, we evaluate the robustness of different methods for point clouds with different similarity transformations. Secondly, we compare the performance of different registration methods for point clouds with different density. Next, we test the robustness of different methods for noisy and defective point clouds. Finally, we show a comprehensive analysis of KSS-ICP, including time cost report and some limitations in practice.

\begin{figure}[]
  \centering
  \includegraphics[width=\linewidth]{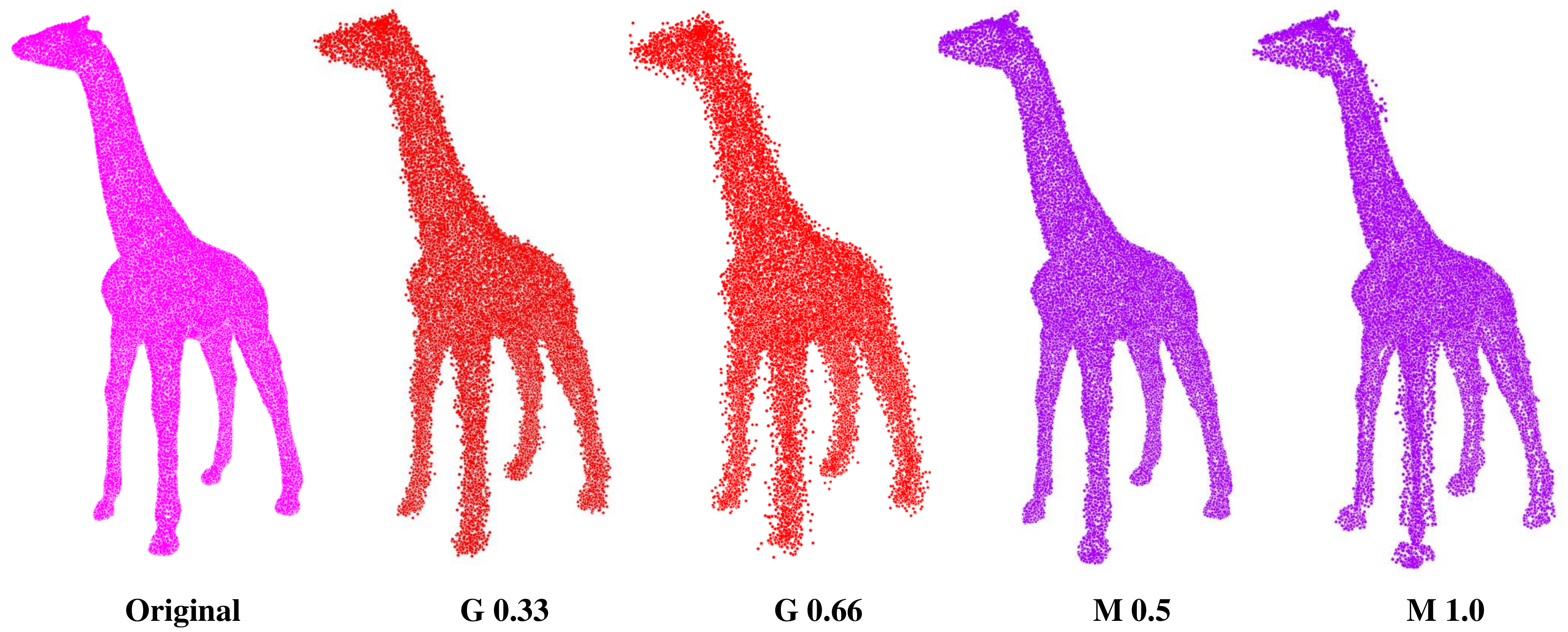}
  \caption{Instances of two kinds of noisy point clouds with different noisy ranges. G and M represent Gaussian and non-zero mean noise; the floating-point numbers represent noisy ranges.}
  \label{fe3}
\end{figure}

\subsection{Robustness for similarity transformations}

It has been discussed that the similarity transformations affect the registration process, especially for different scales and rotations. In this part, we add some similarity transformations with random parameters (scaling$\in\left[0.8, 1.2\right]$, rotation$>30^\circ$ around each axis) into ModelNet40 to construct a source point cloud dataset. Original dataset is used to be the target point cloud dataset. Based on the two datasets, we evaluate the performance of different registration methods, including ICP~\cite{besl1992method}, FPFH~\cite{rusu2009fast}, Go-ICP~\cite{yang2013go}, PonitNetLK~\cite{aoki2019pointnetlk}, PCRNet~\cite{sarode2019pcrnet}, Fast-ICP~\cite{Zhang2022FastICP}, and KSS-ICP. The ICP and FPFH are implemented by the PCL library. The codes of Go-ICP, PonitNetLK, PCRNet, and Fast-ICP are provided by Github (Go-ICP: \href{https://github.com/yangjiaolong/Go-ICP}{\color{black}yangjiaolong/Go-ICP}; PonitNetLK: \href{https://github.com/vinits5/PointNetLK}{\color{black}vinits5/PointNetLK}; PCRNet: \href{https://github.com/vinits5/pcrnet_pytorch}{\color{black}vinits5/pcrnet\_pytorch}; Fast-ICP:\href{https://github.com/yaoyx689/Fast-Robust-ICP}{\color{black}yaoyx689/Fast-Robust-ICP}). In Figure \ref{fe1}, some registration results by different methods are shown. In Tables \ref{ta1}, we compare registration evaluation results by different methods. MSE is the average value of mean squared errors. RMSE is the average value of root mean squared errors. MAE is the average value of mean absolute errors. Euclidean distances of points are computed to generate values of MSE, RMSE, and MAE. Angular measurements of point normal vectors are measured to produce MSE(R), RMSE(R), and MAE(R). The results show that the KSS-ICP achieves more accurate results for point clouds with random similarity transformations.

\begin{figure*}
  \centering
  \includegraphics[width=0.85\linewidth]{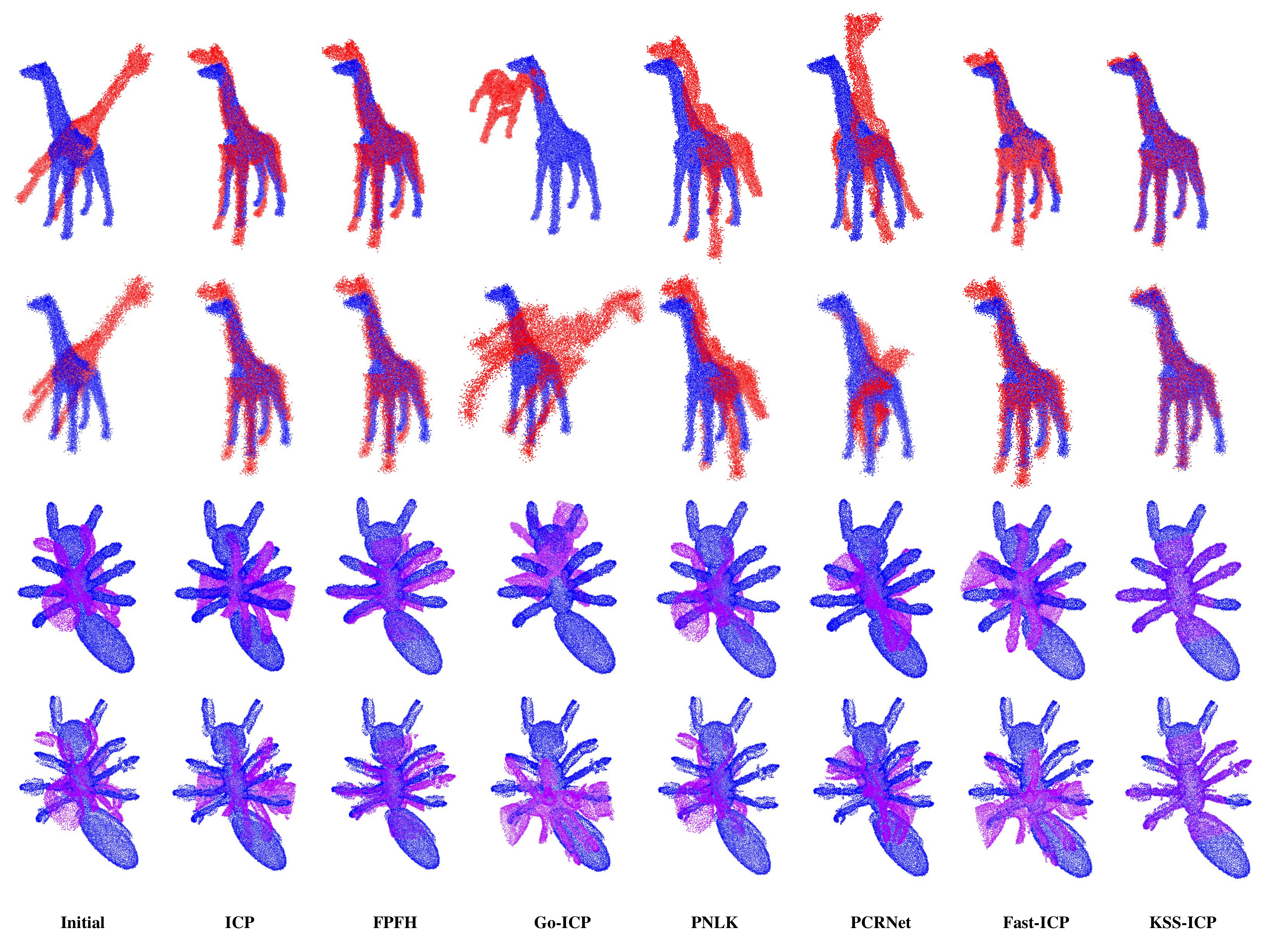}
  \caption{Comparisons of different registration results in SHREC models with Gaussian and non-zero mean noise with different ranges (first row: Gaussian noise with noisy range 0.33; second row: Gaussian noise with noisy range 0.66; third row: non-zero mean noise with noisy range 0.5; fourth row: non-zero mean noise with noisy range 1.0).}
  \label{fe4}
\end{figure*}

\begin{figure*}
  \centering
  \includegraphics[width=0.85\linewidth]{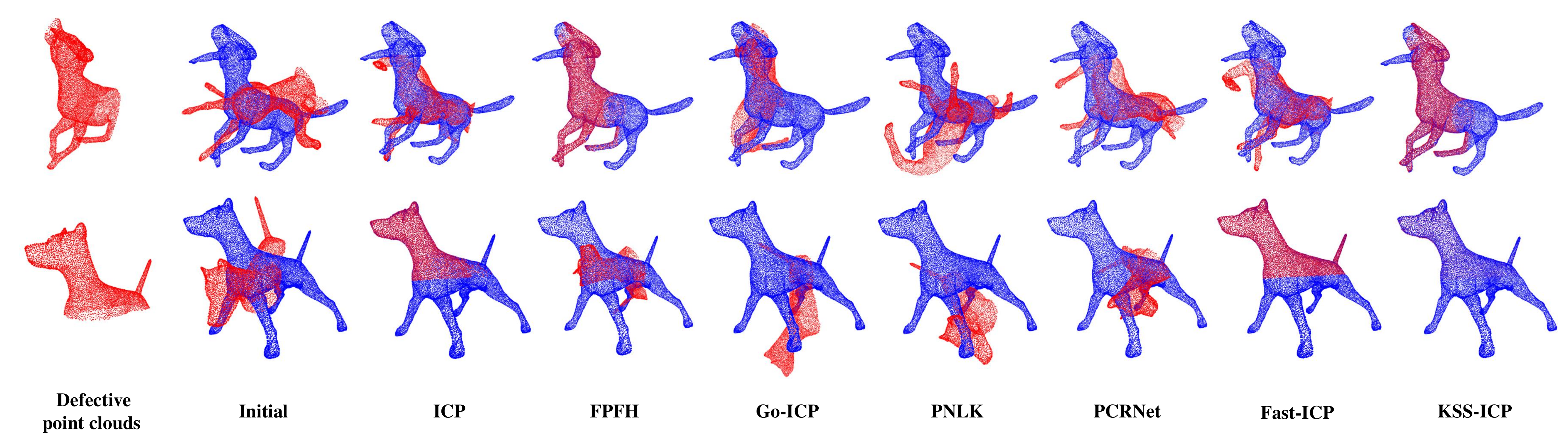}
  \caption{Comparisons of different registration results in SHREC models with defective parts.}
  \label{fe4_1}
\end{figure*}

\subsection{Robustness for different densities}

Different densities of point clouds affect the distance-based metric in the registration process. Therefore the robustness of different densities is important for registration methods. We use a point cloud resample method \cite{huang2009consolidation} to change point densities in RGB-D scenes dataset to construct a source point cloud dataset. Random similarity transformations are also added into the source point cloud dataset with same parameters introduced in previous subsection. Original dataset is used to be the target point cloud dataset. Based on the two datasets, we evaluate the density robustness for different registration methods. In Figure \ref{fe2}, some registration results are shown by different registration methods. Similar to Tables \ref{ta1}, we compare the quantitative results in Tables \ref{ta2}. Benefited from the point cloud representation, KSS-ICP also achieves better robustness for point clouds with different densities.

\begin{table}[]
\caption{Evaluations for different registration methods in SHREC dataset with Gaussian noise. Noisy ranges include 0.33 and 0.66.}
\label{ta3}
\begin{center}
\begin{tabular}{c|c|ccc}
\toprule
\textbf{Noisy Range}                 & \textbf{Method}  & \textbf{MSE} & \textbf{RMSE} & \textbf{MAE} \\\midrule
\multirow{6}{*}{\textbf{0.33}} & \textbf{ICP}     & 2.36E-02     & 1.42E-01      & 1.07E-01     \\
                               & \textbf{FPFH}    & 5.13E-03     & 4.98E-02      & 3.80E-02     \\
                               & \textbf{Go-ICP}  & 4.02E-01     & 5.28E-01      & 4.34E-01     \\
                               & \textbf{PNLK}    & 4.98E-02     & 2.07E-01      & 1.61E-01     \\
                               & \textbf{PCRNet}  & 4.28E-02     & 2.00E-01      & 1.51E-01     \\
                               & \textbf{Fast-ICP}  & 2.91E-02     & 1.54E-01      & 1.01E-01     \\
                               & \textbf{KSS-ICP} & \textbf{1.97E-03}     & \textbf{3.35E-02}      & \textbf{2.63E-02}     \\\midrule
\multirow{6}{*}{\textbf{0.66}} & \textbf{ICP}     & 1.25E-02     & 1.05E-01      & 7.66E-02     \\
                               & \textbf{FPFH}    & 3.19E-03     & 4.48E-02      & 3.34E-02     \\
                               & \textbf{Go-ICP}  & 3.18E-01     & 4.78E-01      & 3.95E-01     \\
                               & \textbf{PNLK}    & 4.60E-02     & 2.04E-01      & 1.58E-01     \\
                               & \textbf{PCRNet}  & 3.75E-02     & 1.87E-01      & 1.36E-01     \\
                               & \textbf{Fast-ICP}  & 2.57E-02     & 1.43E-01      & 9.03E-02     \\
                               & \textbf{KSS-ICP} & \textbf{1.63E-03}     & \textbf{3.28E-02}      & \textbf{2.55E-02} \\ \bottomrule
\end{tabular}
\end{center}
\end{table}

\begin{table}[]
\caption{Evaluations for different registration methods in SHREC dataset with Non-zero mean noise. Noisy ranges include 0.5 and 1.0.}
\label{ta3_2}
\begin{center}
\begin{tabular}{c|c|ccc}
\toprule
\textbf{Noisy Range}                 & \textbf{Method}  & \textbf{MSE} & \textbf{RMSE} & \textbf{MAE} \\\midrule
\multirow{6}{*}{\textbf{0.5}} & \textbf{ICP}     & 2.14E-02     & 1.41E-01      & 1.09E-01     \\
                               & \textbf{FPFH}    & 1.19E-02     & 9.99E-02      & 7.21E-02     \\
                               & \textbf{Go-ICP}  & 1.52E-01     & 3.52E-01      & 2.85E-01     \\
                               & \textbf{PNLK}    & 6.22E-01     & 4.73E-01      & 4.27E-01     \\
                               & \textbf{PCRNet}  & 4.09E-02     & 1.95E-01      & 1.55E-01     \\
                               & \textbf{Fast-ICP}  & 9.65E-03     & 8.37E-02      & 5.67E-02    \\
                               & \textbf{KSS-ICP} & \textbf{7.33E-04}     & \textbf{2.02E-02}      & \textbf{1.63E-02}     \\\midrule
\multirow{6}{*}{\textbf{1.0}} & \textbf{ICP}     & 2.01E-02     & 1.36E-01      & 1.05E-02     \\
                               & \textbf{FPFH}    & 1.12E-02     & 9.65E-02      & 6.91E-02     \\
                               & \textbf{Go-ICP}  & 1.58E-01     & 3.64E-01      & 2.98E-01     \\
                               & \textbf{PNLK}    & 1.25E-01     & 3.01E-01      & 2.56E-01     \\
                               & \textbf{PCRNet}  & 4.03E-02     & 1.93E-01      & 1.53E-01     \\
                               & \textbf{Fast-ICP}  & 9.38E-03     & 8.41E-02      & 5.72E-02     \\
                               & \textbf{KSS-ICP} & \textbf{7.89E-04}     & \textbf{2.12E-02}      & \textbf{1.68E-02} \\ \bottomrule
\end{tabular}
\end{center}
\end{table}

\begin{table}[]
\caption{Evaluations for different registration methods in SHREC partial dataset.}
\scriptsize
\label{ta4}
\begin{center}
\resizebox{\columnwidth}{!}{
\begin{tabular}{c|cccccc}
\toprule
\textbf{Method}  & \textbf{MSE}     & \textbf{RMSE}    & \textbf{MAE}     & \textbf{MSE(R)}  & \textbf{RMSE(R)} & \textbf{MAE(R)}  \\\midrule
\textbf{ICP}     & 1.91E-3          & 3.01E-2          & 2.34E-2          & 4.35E-1          & 4.88E-1          & 4.26E-1          \\
\textbf{FPFH}    & 2.32E-3          & 2.74E-2          & 2.12E-2          & 3.28E-1          & 3.75E-1          & 3.23E-1          \\
\textbf{Go-ICP}  & 1.71E-1          & 3.29E-1          & 2.68E-1          & 1.07E-0          & 1.02E-0          & 9.51E-1          \\
\textbf{PNLK}    & 2.59E-2          & 1.41E-1          & 1.12E-1          & 9.71E-1          & 9.72E-1          & 8.81E-1          \\
\textbf{PCRNet}  & 3.13E-2          & 1.69E-1          & 1.28E-1          & 1.15E-0          & 1.07E-0          & 1.01E-0          \\
\textbf{Fast-ICP} & 1.01E-2 & 8.58E-2 & 5.88E-2 & 8.03E-1 & 8.91E-1 & 7.85E-1\\
\textbf{KSS-ICP} & \textbf{9.32E-4} & \textbf{2.14E-2} & \textbf{1.67E-2} & \textbf{3.15E-1} & \textbf{3.49E-1} & \textbf{3.05E-1}\\\bottomrule
\end{tabular}}
\end{center}
\end{table}

\subsection{Robustness for noisy and defective point clouds}

\begin{figure*}
  \centering
  \includegraphics[width=0.95\linewidth]{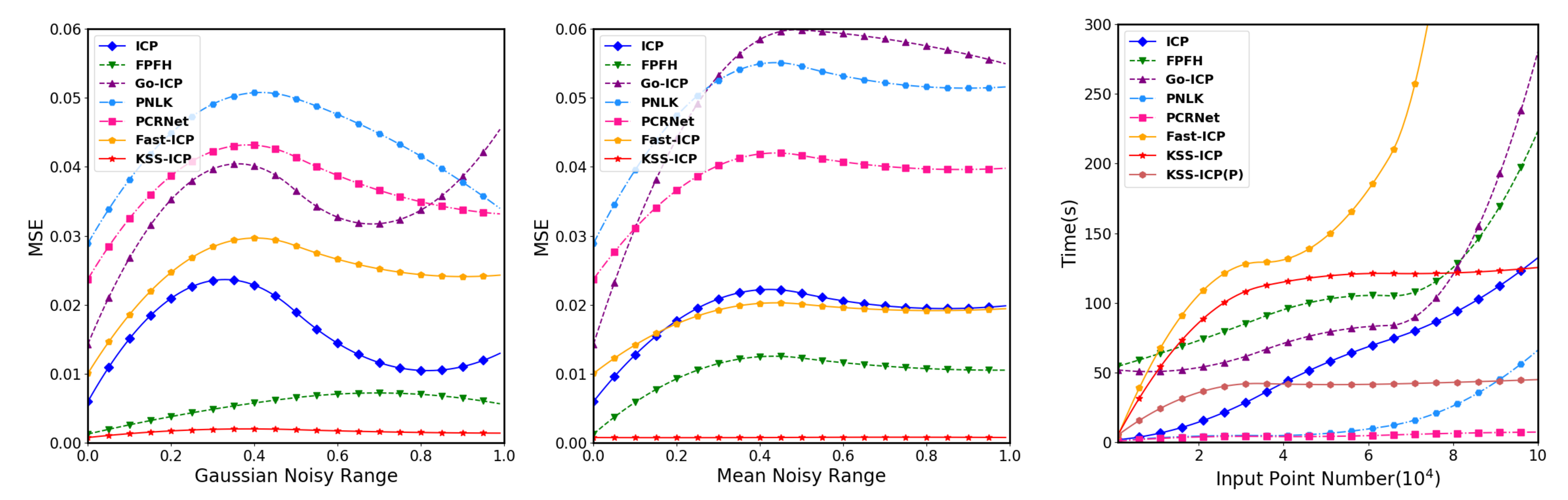}
  \caption{MSE (Euclidean distance) and time cost curves generated by different methods in SHREC dataset. First chart: MSE curves for point clouds with Gaussian noise; second chart: MSE curves for point clouds with non-zero mean noise; third chart: time cost report.}
  \label{fe5}
\end{figure*}

As mentioned before, noise and defective parts in point clouds can not be avoided in wild scanning. The robustness for noisy and defective point clouds should be measured for registration methods. We build five test sets from SHREC to evaluate the robustness. Two kinds of noise are added into SHREC models to generate four test sets with different noisy ranges. One kind of noise is Gaussian noise generated by
\begin{equation}
\label{e_noise}
\begin{array}{c}
p_i'=p_i+n_i\cdot m_i,\;\\
m_i\in\{m\},\{m\}\sim N(0,\sigma^2),\sigma=r\ast l_k,
\end{array}
\end{equation}
where $p_i’$ is the generated noise that is computed from the original point $p_i$ with a random movement $m_i$ according to the normal vector $n_i$. The values of $\{m\}$ satisfy the normal distribution. The $\sigma$ is the distributed control parameter that is computed by the input noisy range $r$ and $l_k$ ($l_k$ is the average length between points and their $k$ neighbors, $k = 12$ be default). The range $r$ controls the value of $\sigma$ that reflects the degree of the noise. The other kind of noise is non-zero mean noise. We just change the normal distribution to uniform one based on Equation~\ref{e_noise} to generate the noise, $\{m\}\sim U(0,\sigma^2)$. In Figure~\ref{fe3}, some instances of noisy point clouds are shown. In Figure \ref{fe4}, some registration results achieved from the test sets are shown. In Tables \ref{ta3} and \ref{ta3_2}, we compare the quantitative results from different methods based on two kinds of noise. Based on the visualization and quantitative analysis, our method achieves more accurate results. In Figure \ref{fe5} (two sub-pictures on the left), curves of MSE values from registration results with different kinds of noise are shown. The curves reflect the noisy sensitivity of different registration methods. It is clear that our method has better performance of noisy robustness.

For evaluation of defective parts robustness, the artificial deletions are processed into SHREC to generate the fifth test set. Based on the set, we compare the registration results from different methods. In Figure \ref{fe4_1}, two instances are shown. In Table \ref{ta4}, the quantitative results of different method are compared. Based on the test results, it can be proved that the KSS-ICP achieves better robustness to noisy and defective point clouds.

\subsection{Further Analysis}

Based on the experimental data, a comprehensive analysis is shown in this part. The complexity of ICP is $O(N^2)$ that is higher in registration task. The FPFH improves the complexity to the $O(Nk)$, $k$ is the specified neighbor number. However, the time cost of the FPFH is larger than ICP for same point clouds. The reason is that the computation cost of FPFH feature extraction is huge. The PointNetLK achieves improvement for registration with $O(N)$ complexity. In fact, the point number $N$ in PointNetLK is controlled by the network structure. The computation cost is produced by the pre-processing, which is same to the PCRNet. For Go-ICP and Fast-ICP, the complexities are improved to approach $O(N)$. However, the convergence performance of the two methods is not stable, especially for point clouds with different scales. The complexity of KSS-ICP is $O(N|C||O|)$, $|C|$ and $|O|$ represent the numbers of elements in $\{C\}$ and $\{O\}$ discussed in Equation~\ref{e13}. With the parallel acceleration, $O(N|C||O|)$ can be approximately improved to $O(N)$. In Figure \ref{fe5} (the third sub-picture), time cost curves of different registration methods are shown. KSS-ICP(P) means the KSS-ICP with the GPU-based parallel acceleration. Based on the curves, it can be seen that the deep learning-based methods are faster.

\begin{table}[]
\caption{Time cost reports of point clouds with different scales (k: 1000 points) for different methods. Some results that the time cost is more than 300 second are ignored. KSS-ICP(P): KSS-ICP with parallel acceleration.}
\label{ta5}
\resizebox{\columnwidth}{!}{
\begin{tabular}{c|cccccc}
\toprule
\textbf{Method}     & \textbf{5k}    & \textbf{10k}  & \textbf{50k}  & \textbf{100k} & \textbf{500k}  & \textbf{1,000k} \\\midrule
\textbf{ICP}        & 1.9s           & 15.7s         & 58.8s        & 139.2s        & ——             & ——              \\
\textbf{FPFH}       & 55.5s          & 75.5s         & 102.1s        & 223.1s        & ——             & ——              \\
\textbf{Go-ICP}     & 49.8s          & 51.3s         & 57.0s         & ——            & ——             & ——              \\
\textbf{PNLK}       & \textbf{0.3s} & 4.1s          & 4.2s          & 66.1s          & ——             & ——              \\
\textbf{PCRNet}     & 1.0s          & \textbf{3.2s} & \textbf{4.3s} & \textbf{5.2s} & ——             & ——              \\
\textbf{Fast-ICP}     & 4.5s          & 88.6s & 126.6s & —— & ——      & ——             \\
\textbf{KSS-ICP}    & 6.5s           & 85.7s         & 119.2s        & 125.4s         & 282.2s         & ——              \\
\textbf{KSS-ICP(P)} & 6.1s           & 37.9s         & 41.3s         & 44.9s         & \textbf{79.9s} & \textbf{124.3s}\\ \bottomrule
\end{tabular}}
\end{table}

For the performance of MSE-based evaluation, the deep learning-based methods do not achieve accurate MSE values in test datasets, especially for the point clouds with different scales and large rotations. The influence of similarity transformation is not reduced during the encoding of the methods. It takes some unstable factors for registration. The FPFH achieves accuracy registration results in different conditions. Although it cannot reduce the influence of different scales, the rotation alignment is correct in most cases.
However, the performance of the FPFH is affected by the noise and defective parts. The limitations of Go-ICP are sensitive to noisy and defective point clouds, and its convergence speed is slow for point clouds with different scales. For point clouds with large number ($>500k$), the performance of most methods is decreased. There have two reasons: the time cost of pre-processing is increased; the convergence speed of the optimization process is changed to slow. In Table \ref{ta5}, we show the time cost reports for different methods in point clouds with large number ranges($[5k-1,000k]$). Benefited from the point cloud representation in and the proposed alignment, KSS-ICP can register point clouds with large number efficiently. With the parallel acceleration, it can be further improved.

\begin{figure}[t]
  \centering
  \includegraphics[width=0.95\linewidth]{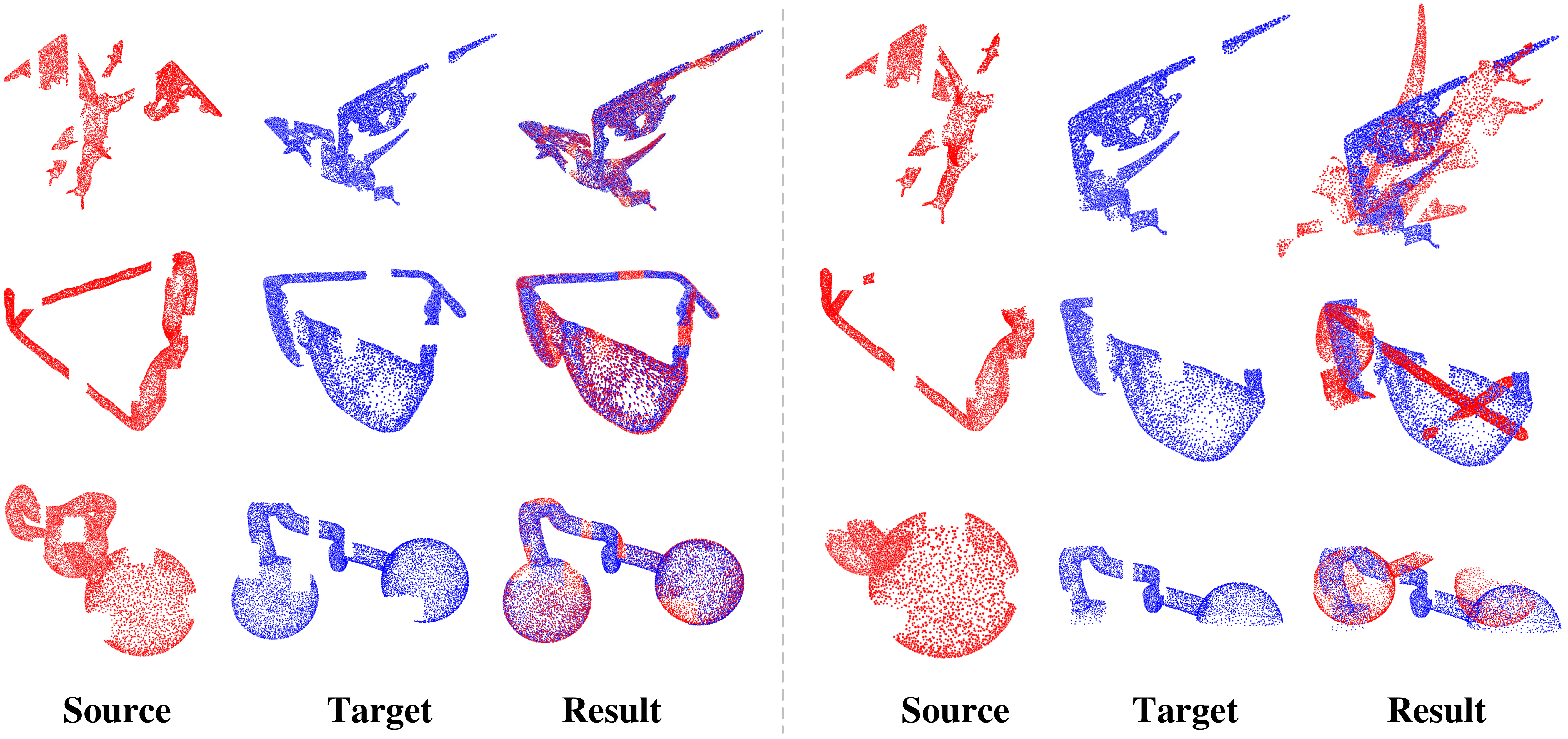}
  \caption{Some instances of partial-partial registration by our method. Bounded by grey dotted line, left: positive instances; right: negative instances.}
  \label{fe7}
\end{figure}

Although the KSS-ICP has many advantages for various registration tasks, there are some limitations still exist in practice. Firstly, our method searches the potential alignment in global view which increases the computation. As mentioned before, the time cost of KSS-ICP is not faster than deep learning frameworks in some cases where the point cloud number is smaller than 100k. The performance depends on the parallel acceleration that is limited by the hardware. Secondly, the function of KSS-ICP for partial-global is restricted. KSS-ICP is sensitive to point clouds with an incomplete geometric structure that smaller than half of the global one. In Figure~\ref{fe7}, some instances are used to illustrate the limitation. The performance of the method is reduced when the global structure is broken. The reduction produces error registration results like negative instances in Figure~\ref{fe7}. The reason is that the alignment of the KSS-ICP is implemented based on the global shape matching. Fortunately, when the missing parts are not affecting the representation of global shape feature (the centers can be aligned by our exhaustive searching strategy), our method still works well. Some positive instances are shown in Figure~\ref{fe7} at the same time. Considering the balance between the efficiency and feasibility, our method can still be regarded as a reasonable and practical solution for registration.

\section{Conclusions}

We have proposed a point cloud registration method, KSS-ICP, based on Kendall shape space (KSS) theory, as a feasible solution to avoid the difficulty in registration. The point cloud representation in KSS reduces influences of different point densities, locations, and scales. Combining the alignment with parallel acceleration and ICP method, KSS-ICP achieves registration results for point clouds with acceptable complexity. It does not require complex geometric feature analysis and optimization. With a concise implementation, KSS-ICP avoids the local optima as much as possible while improving the noise robustness. Benefited from the parallel acceleration, the performance is not reduced when the search range is expanded in KSS. A large number of computation units of GPU search candidate results at the same time without additional time cost. It also supports partial-global registration for point cloud with defective parts. Experiments show that the KSS-ICP has good performance for registration with different conditions. Comparing to the traditional methods, it achieves a better balance between accuracy and efficiency.

In future work, we consider combining the KSS-based measurement and point cloud-based deep network to implement registration. We expect to design a local shape alignment based on the deep encoding method to remove the limitation in partial-global registration task.

\ifCLASSOPTIONcaptionsoff
  \newpage
\fi
\bibliographystyle{IEEEtran}
\bibliography{IEEEabrv,IEEEBib}

\end{document}